\documentclass{article} %
\usepackage{iclr2023_conference,times}

\usepackage{amsmath,amsfonts,bm}

\def\Figref#1{Figure~\ref{#1}}

\def\Secref#1{Section~\ref{#1}}

\def\eqref#1{equation~\ref{#1}}
\def\Eqref#1{Equation~\ref{#1}}

\def\Algref#1{Algorithm~\ref{#1}}

\def\1{\bm{1}}

\DeclareMathAlphabet{\mathsfit}{\encodingdefault}{\sfdefault}{m}{sl}
\SetMathAlphabet{\mathsfit}{bold}{\encodingdefault}{\sfdefault}{bx}{n}

\def\gD{{\mathcal{D}}}

\def\gG{{\mathcal{G}}}

\def\gL{{\mathcal{L}}}

\def\gS{{\mathcal{S}}}

\def\gU{{\mathcal{U}}}

\def\gX{{\mathcal{X}}}
\def\gY{{\mathcal{Y}}}

\DeclareMathOperator*{\argmax}{arg\,max}
\DeclareMathOperator*{\argmin}{arg\,min}

\usepackage{hyperref}
\usepackage{url}

\usepackage{url}
\usepackage{hyperref}
\usepackage{amsmath}
\usepackage{booktabs}
\usepackage{xspace}
\usepackage{pifont}
\usepackage{threeparttable}
\usepackage{tikz}
\usepackage{graphicx} 
\usepackage{pgfplots}
\usepackage{multirow}
\RequirePackage{algorithm}
\RequirePackage{algorithmic}
\usepackage{enumitem}
\usepackage{wrapfig}
\usepackage{subcaption}
\usepackage[normalem]{ulem}
\usepackage{enumitem}

\newcommand{\adaicl}{\textsc{AdaIcl}\xspace}
\newcommand{\adaiclp}{\textsc{AdaIcl+}\xspace}
\newcommand{\adaiclb}{\textsc{AdaIcl}-base\xspace}
\newcommand{\mcp}{\textsc{MaxCover}\xspace}

\title{Which Examples to Annotate for In-Context Learning? Towards Effective and Efficient Selection}

\author{Costas Mavromatis$^\dag$\thanks{Work done while interning at Amazon Web Services. $^\dag$Corresponding authors.} \\
University of Minnesota \\
\texttt{mavro016@umn.edu} \\
\And
Balasubramaniam Srinivasan$^\dag$ \\
Amazon Web Services \\
\texttt{srbalasu@amazon.com} \\
\And
Zhengyuan Shen \\
Amazon Web Services \\
\texttt{donshen@amazon.com} \\
\And
Jiani Zhang \\
Amazon Web Services \\
\texttt{zhajiani@amazon.com} \\
\And
Huzefa Rangwala \\
Amazon Web Services \\
\texttt{rhuzefa@amazon.com} \\
\And
Christos Faloutsos \\
Amazon Web Services \\
\texttt{faloutso@amazon.com} \\
\And
George Karypis \\
Amazon Web Services \\
\texttt{gkarypis@amazon.com} \\
}

\iclrfinalcopy %
\begin{document}

\maketitle

\begin{abstract}
 Large Language Models (LLMs) can adapt to new tasks via in-context learning (ICL). ICL is efficient as it does not require any parameter updates to the trained LLM, but only few annotated examples as input for the LLM. In this work, we investigate an active learning approach for ICL, where there is a limited budget for annotating examples. We propose a model-adaptive optimization-free algorithm, termed \adaicl, which identifies examples that the model is uncertain about, and performs semantic diversity-based example selection. Diversity-based sampling improves overall effectiveness, while uncertainty sampling improves budget efficiency and helps the LLM learn new information. Moreover, \adaicl poses its sampling strategy as a Maximum Coverage problem, that dynamically adapts based on the model's feedback and can be approximately solved via greedy algorithms. Extensive experiments on nine datasets and seven LLMs show that \adaicl improves performance by 4.4\% accuracy points over SOTA (7.7\% relative improvement), is up to $3\times$ more budget-efficient than performing annotations uniformly at random, while it outperforms SOTA with $2 \times$ fewer ICL examples. Our code is available at \url{ https://github.com/amazon-science/adaptive-in-context-learning}.

\end{abstract}

\section{Introduction} \label{sec:intro}

Large Language Models (LLMs) have shown remarkable performance in various natural language tasks.  One of the LLMs' advantages is their ability to perform few-shot learning~\citep{brown2020fewshot}, where they can adapt to new tasks, e.g., topic classification or sentiment prediction, via in-context learning (ICL).  ICL uses few-shot labeled examples $\{(x_i, y_i)\}_{i=1}^k$, e.g., $(x_i, y_i)=$ \texttt{\footnotesize (Amazing movie!, positive)}, to construct a prompt $P$.
Prompt $P$ is used as a new input to the LLM, e.g., 
``\texttt{\footnotesize Amazing movie!: positive \textbackslash n Awful acting: negative \textbackslash n Terrible movie:}'', before making predictions for the query $x_{\text{test}}$. 
The new input enables the LLM to infer the missing label $y_{\text{test}}$ by conditioning the generation on the few-shot examples. As semantically similar demonstrations to the test query improve ICL performance~\citep{liu2021makes}, it is common practice that a $k$-NN retriever is used to determine the $k$-nearest examples for each test query.

ICL is efficient as it does not require any parameter updates or fine-tuning, wherein users can leverage ICL to generate task-adaptive responses from black-box LLMs.
However, ICL is sensitive to the input prompt~\citep{lu2021fantastically} (the art of constructing successful prompts is referred to as prompt engineering),  where acquiring ground-truth labeling of the input demonstrations is important
for good ICL performance~\citep{yoo2022ground}. Ground-truth labeling requires expert annotators and can be costly, especially for tasks in which the annotators need to provide elaborate responses~\citep{wei2022chain}. Apart from lowering the labeling cost, carefully reducing the number of the ICL examples can benefit inference costs and the LLM's input context length requirements.
Consequently, we study the following problem: \textit{Given a budget $B$, which examples do we select to annotate and include in the prompt of ICL?}

Given an unlabeled set (pool) where we can draw ICL examples from, the  above selection problem resembles a typical active learning setting~\cite{settles2009active,zhang-etal-2022-survey}.
Active learning selects \textit{informative} examples, e.g., via uncertainty sampling~\citep{Lewis1994ASA}, which are used to improve the model's performance. 
Although traditional active learning involves model parameter updates, uncertainty sampling has been explored in an optimization-free manner for ICL with black-box LLMs~\citep{diao2023active,pitis2023boosted}. Yet, recent studies show that  uncertainty sampling yields inferior performance in comparison to other approaches~\citep{margatina2023active} and thus, current methods rely on semantic diversity to determine which examples are the most informative. For example, AutoCoT~\citep{zhang2022automatic} performs clustering based on semantic similarity and selects the most representative examples of each cluster. In order to adapt the selection based on the LLM used,  Vote$k$~\citep{su2022selective} selects diverse examples with respect to the LLM's feedback, i.e., examples that the model is both uncertain and confident about.  However, these approaches do not consider which examples help the LLM learn new information and may waste resources for annotating examples whose answers are already within the model's knowledge.

\begin{wrapfigure}{r}{0.5\textwidth} %
    \centering
    \vspace{-0.2in}
    \includegraphics[width=0.4\textwidth]{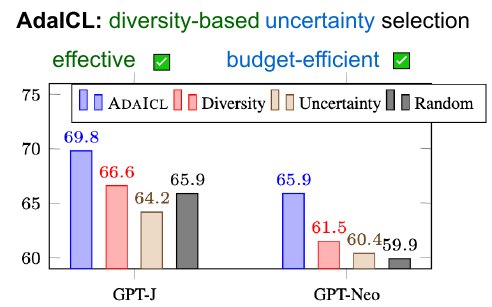}
    \vspace{-0.1in}
    \caption{\adaicl effectively combines diversity and uncertainty sampling, outperforming other strategies in the low-resource scenario, averaged over seven datasets. Here, the budget is 20 annotations for retrieval-based 5-shot ICL.}
    \vspace{-0.17in}
    \label{fig:ada_intro}
\end{wrapfigure}
  
To overcome the aforementioned limitations of active learning for ICL, we pair uncertainty-based sampling with diversity-based sampling.  
To best combine the two sampling strategies, we propose a model-adaptive optimization-free method, termed \adaicl, which is tailored to retrieval-based $k$-shot
ICL. \adaicl uses the LLM's feedback (output probabilities) to identify the examples that the  model is most uncertain about (\emph{hard examples}). The algorithm then identifies different semantic regions of hard examples, with the goal to select the most representative examples within each region. Diversity-based sampling is posed as the well-studied Maximum Coverage problem (\mcp), which can be approximately solved via greedy algorithms.

By selecting representative examples of diverse hard regions, \adaicl aims for \emph{effectiveness}, so that it helps the LLM learn information that it does not already know. Moreover,  \adaicl is \emph{efficient} and results in budget savings by omitting the selection of examples that the model already knows how to tackle (easy examples). Finally, we show that \adaicl's uncertainty sampling improves the LLM's calibration~\citep{jiang2021can, zhao2021calibrate}, i.e., how model's confidence aligns with epistemic uncertainty, which measures how well the model understands the task.

We conduct experiments on nine datasets across five NLP tasks (topic classification, sentiment analysis, natural language inference, summarization, and math reasoning) with seven LLMs of varying sizes (1.3B to 65B parameters), including LLMs such as GPT-J~\citep{wang2021gpt-j}, Mosaic~\citep{MosaicML2023Introducing}, Falcon~\citep{penedo2023falcon}, and LLaMa~\citep{touvron2023llama}. Experimental results show that (i) \adaicl is effective with an average performance improvement of 4.4\% accuracy points over SOTA (\Figref{fig:ada_intro}),  (ii) \adaicl is robust and achieves up to $3\times$ budget savings over random annotation, while it needs $2 \times$ fewer ICL examples than SOTA (\Secref{sec:res2}), and (iii) \adaicl improves the calibration of the LLM's predictions (\Secref{sec:res-calib}).

\section{Related Work} \label{sec:related}

\textbf{ICL Mechanism}. In-context learning (ICL), also referred to as few-shot learning~\citep{brown2020fewshot}, has been shown to elicit reasoning capabilities of LLMs without any fine-tuning~\citep{wei2022emergent,bommasani2021opportunities,dong2022survey-icl}.
While ICL performance correlates to the pretraining data distribution~\citep{xie2021explanation,hahn2023theory,shin2022effect,razeghi2022impact,chan2022data} and improves with larger LMs~\citep{bansal2022rethinking,wei2023larger}, recent works study how the transformer architecture~\citep{vaswani2017attention} enables in-context learning~\citep{akyurek2022learning,olsson2022context}. Theoretical analyses and empirical studies show that ICL is a learning algorithm that acts as a linear regression~\citep{akyurek2022learning,garg2022icl-linear,von2023transformers,zhang2023trained} and ridge (kernel) regression~\citep{han2023icl-kernel,bai2023statisticians} classifier. In this work, we leverage the recent connections of ICL with kernel regression to highlight the challenges of active learning for ICL.

\textbf{Prompt Influence for ICL}. Although ICL is widely used with LLMs, its success is not guaranteed as it is sensitive to the input prompts~\citep{lu2021fantastically,chen2022relation}. ICL tuning improves stability~\cite{min2021metaicl,chen2021meta,xu2023knn} but requires additional training data, while other works analyze how the prompt design and the semantics of the labels~\cite{min2022rethinking,yoo2022ground,wang2022towards,wei2023larger} affect ICL performance. More related to our setting, \citet{liu2021makes},\citet{Rubin2021LearningTR}, \citet{margatina2023active}, and \citet{su2022selective} illustrate the importance of the $k$-NN retriever for ICL. Our work also highlights the importance of the $k$-NN retriever and provides new insights for improving ICL performance.

\textbf{Active Learning for NLP}. Active learning~\citep{settles2009active} for NLP has been well-studied~\citep{zhang-etal-2022-survey} with applications to text classification~\citep{schröder2020survey}, machine translation~\citep{haffari2009active}, and name entity recognition~\citet{erdmann-etal-2019-practical}, among others. \citet{ein-dor-etal-2020-active} studied the application of traditional active learning techniques~\citep{Lewis1994ASA,sener2018active} for BERT pretrained models~\citep{devlin2019bert}, with many works following up~\citep{margatina-etal-2021-active, schroder2021revisiting}. However, these approaches rely on LM fine-tuning, which can be more difficult to scale for larger models with tens of billions of parameters. Furthermore, most of the current approaches of active learning for ICL~\citep{zhang2022activeICL,li2023finding,nguyen2023context,shum2023automatic,ma2023fairness} assume a high-resource setting, where a large set of ICL examples is already annotated.  This limits the applicability in practical low-resource scenarios~\citep{perez2021true}, where annotations are costly to obtain.

\section{Problem Statement \& Motivation} \label{sec:challenges}

Given an unlabeled set $\gU = \{x_i\}_{i=1}^N$ and a fixed budget $B \in \mathbb{Z}^+$, the goal is to select a subset $\gL \subset \gU$, where $\gL = \{ (x_i, y_i)\}_{i=1}^B$ contains $B$ selected examples that are annotated.
 Due to token-length limits or inference cost considerations, we consider a $k$-shot ICL inference, where 
set $\gL$ is used to draw $k$-shot ICL examples from ($k < B$). The $k$-shot examples are used to construct a new prompt $P$ as input to the LLM by
\begin{equation}
P = \pi (x_1, y_1) \oplus \pi (x_2, y_2)  \oplus \cdots  \oplus \pi(x_k, y_k) \oplus \pi(x_\text{test}, *).
\end{equation}
Template $\pi$ denotes a natural language verbalization for each demonstration $(x,y)$ and it also expresses how the labels $y$ map to the target tokens.
The selected $B$ examples for set $\gL$ should maximize the ICL model performance on the testing set.

\begin{figure*}[t]
    \centering
    \includegraphics[width=1.0\linewidth]{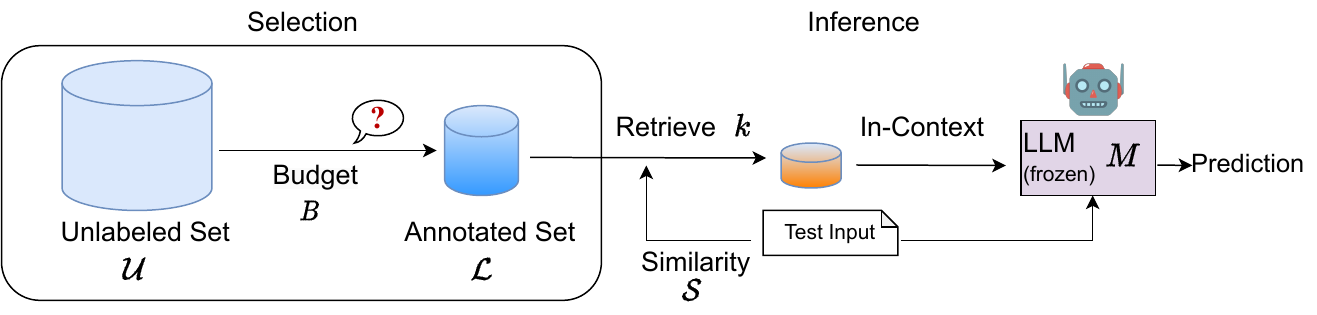}
    \caption{Our studied problem setting. Given an unlabeled set $\gU$ and a fixed budget $B$, the goal is to select the $B$ most informative examples for annotation (set $\gL$), which are used to maximize ICL performance with an LLM $M$.  During ICL inference, a $k$-NN retriever based on a similarity space $\gS$ determines the $k$-shot demonstrations for each test instance.  
    }
    \label{fig:setting}
\end{figure*}

To determine which ICL examples to use (and in which order), a $k$-NN retriever selects the top-$k$ examples from $\gL$, e.g., $(x_k, y_k)$, for a test instance $x_{\text{test}}$ based on the similarity between $x_i \in \gL$ and $x_{\text{test}}$ over a semantic space $\gS$ (the order is determined by similarity scores). Employing a $k$-NN retriever for ICL example selection has demonstrated superior performance over random or fixed example selection ~\citep{su2022selective,margatina2023active,liu2021makes}. We illustrate the overall problem setting in \Figref{fig:setting}. %

To understand the impact of the ICL examples on model predictions, we express ICL inference as a non-parametric kernel regression, following the theoretical works from~\citet{han2023icl-kernel, bai2023statisticians}. The prediction for the test instance $x_{\text{test}}$ is related to
\begin{equation}
    \tilde{y}_{\text{test}} = \frac{\sum_{i=1}^k y_i K_\gD (x_{\text{test}}, x_i)}{\sum_{i=1}^k K_\gD (x_{\text{test}},x_i)},
    \label{eq:kernel}
\end{equation}
where $K_\gD(x_{\text{test}}, x_i)$ is a kernel that measures the similarity between  $x_{\text{test}}$ with each of the $k$-shot retrieved instance $x_i$, which depends on the pretraining data distribution $\gD$ for model $M$. 
The importance of the $k$-NN retriever in the lens of \Eqref{eq:kernel} becomes clear: it fetches \emph{semantically similar} examples to compute  $\tilde{y}_{\text{test}}$ by regressing over their labels $y_i$. However, it does not account for which examples help the model learn the task to a larger extent, which depends on $K_\gD$ and is infeasible to determine for pre-trained LLMs as there is usually no direct access to $\gD$.

\section{Adaptive Example Annotation for ICL} \label{sec:adaicl}

{\em Can we improve example selection for ICL motivated by the aforementioned challenges?} We answer affirmatively by putting forth a \textbf{diversity-based uncertainty sampling} strategy, termed \adaicl, which adaptively identifies semantically different examples that help the model learn new information about the task.

\begin{figure*}[t]
    \centering
    \includegraphics[width=1.0\linewidth]{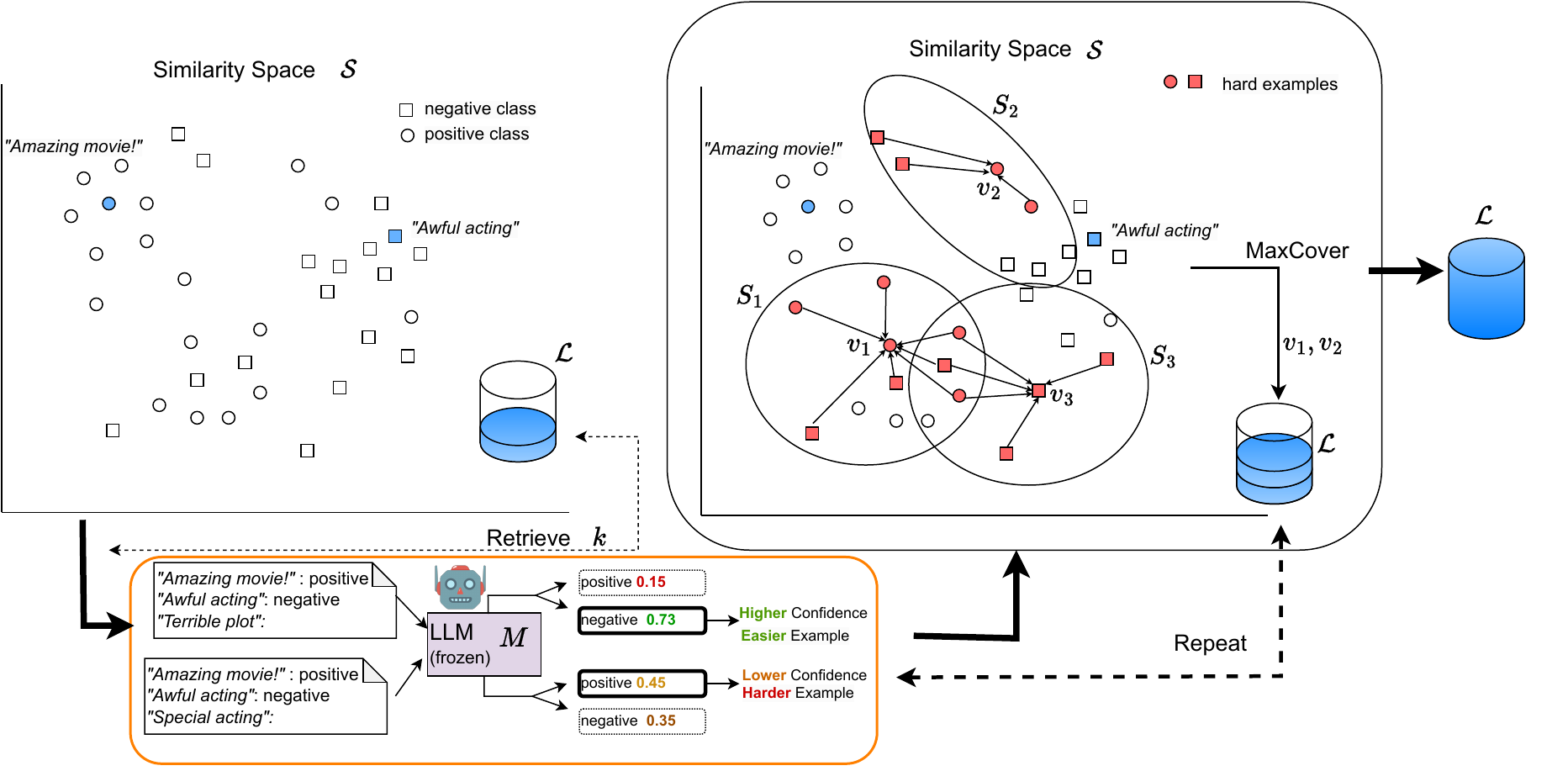}
    \caption{\adaicl algorithm. \adaicl uses $k$-shot ICL to determine which examples the model $M$ is uncertain for (hard examples). Then, it performs diversity-based uncertainty sampling over $\gS$ by optimizing the \mcp problem in \Eqref{eq:mcp-1} via \Algref{alg:greedy} to identify the examples that help the model learn new information. The process is repeated until the budget $B$ is exhausted, and when done, it returns the annotated set $\gL$.}
    \label{fig:framework}
\end{figure*}

The \adaicl algorithm works as follows (see \Figref{fig:framework} for an overview). First, it uses the LLM's feedback to determine which examples the model is most uncertain about, which we refer to as \emph{hard examples}. Then, it builds different regions of hard examples and the goal is to select the most representative example for each region. The regions are determined based on the semantic similarity space $\gS$, apart from the LLM's feedback, and denser regions are considered as more important.

By selecting representative examples of dense hard regions, we seek \textbf{effectiveness} by assuming that (i) these examples are informative for the hard examples that are close in the semantic space, and (ii) these examples will likely be retrieved during inference to improve predictions for similar hard examples (\Eqref{eq:kernel}).
Moreover, by focusing on hard examples, we are interested in examples that the model is uncertain about in order to fill two needs. First, we seek \textbf{budget-efficiency} by avoiding annotating examples that the model already knows how to tackle, i.e., if $K_\gD(x_{\text{test}}, \cdot)$ is high. Second, we seek examples that the model is \emph{truly uncertain} for and can help the model learn the task.

\subsection{\adaiclb: A $k$means Approach} \label{sec:base}

A straightforward solution is to perform $k$means clustering~\citep{macqueen1967kmeans} over the identified hard examples for the model, where different clusters represent regions of hard examples.

To determine the set $\gU_h$ of hard examples, we use the LLM's feedback as follows. We assume we are given a small initially annotated pool $\gL_0$ for $k$-shot ICL (if $\gL_0 = \emptyset$, we perform zero-shot ICL) and use the LLM $M$ to generate a prediction for each $x_i \in \gU$. For classification problems, we compute the
conditional probability of each class $y \in \gY$, and the label $\tilde{y}_i$ with the maximum conditional probability is returned as the prediction for $x_i$. Then, the probability score of the predicted label is regarded as the model's confidence score $u_i$, where lower probability means that the model is less certain for its prediction $\tilde{y}_i$~\citep{min2021metaicl}. For generation problems, we average the log-probabilities of the generated tokens as the model's confidence score $u_i$.  
We sort the examples $x_i \in \gU$ based on their uncertainty scores $u_i$, and select the top-$N_\theta$ out of $N$ total examples, which are collected to $\gU_h$. Here, $N_\theta = \lfloor \theta N \rfloor$ and $\theta \in [0, 1]$ is a hyperparameter  with default value $\theta = 0.5$, which is the portion of the examples that we consider as hard ones.

Then, we can select representative examples for each region by sampling the example closest to each of the cluster centroids. Here, the number of clusters for $k$means is $B$, so we sample as many examples as the budget $B$ allows.
We refer to that approach as \adaiclb and its algorithm is summarized in Appendix~\ref{app:adaiclbase}. Yet, \adaiclb suffers from the known limitations of $k$means clustering. It is sensitive to outlier examples, which may be selected to be annotated, under the assumption that the $B$ regions formed by $k$means are equally important, and it does not account for the effect of overlapping clusters. We provide a failing case that the selected examples may not help the model understand the task in Appendix~\ref{app:adaiclbase}.

\subsection{\adaicl: Selection by Maximum Coverage } \label{sec:sub-adaicl}

\adaicl overcomes the aforementioned limitations of \adaiclb by quantifying whether each each example can convey new information to the model. \adaicl constructs semantic regions $S_i$ \emph{around every} hard example $x_i$ and solves a Maximum Coverage (\mcp) problem that accounts for information overlaps between different regions. \mcp aims to select regions that cover the hardest examples, giving  importance to denser regions and disregarding regions already covered (see \Figref{fig:framework} for an overview).

Formally, \mcp takes $m$ sets $\{S_1, \dots, S_m\}$ (regions that contain semantically similar examples) and a number $B$ as input. Each set includes some examples, e.g., $S_i=\{x_1,x_2, \dots ,x_n\}$ and different sets may have common examples, while the goal is to select the $B$ most representative sets that include (cover) as many examples as possible. We assume that if an example is marked as covered by another selected set, it conveys little new information to the model. 
In our setting, we are interested in hard examples, which we collect in the set $\gU_h$ as previously explained. First, we discuss how we construct the regions and then we provide the \mcp problem.

\textbf{Set Construction.} \label{sec:sets}
We represent the region $S_i$ around each example $x_i$ as its egonet. Initially, we build a global graph $\gG_m$ %
as the $m$-nearest neighbors graph. The nearest neighbors are   determined based on a semantic space $\gS$ given by off-the-self encoders, such as SBERT~\citep{reimers2019sentence}. We compute SBERT embeddings of each query $x_i$ and determine its neighbors based on cosine similarity of the embeddings.
Graph $\gG_m$ depends on the similarity space, while 
deriving sets $S_i$ from the global graph $\gG_m$ depends on the LLM's feedback as we are interested in hard examples $x_i \in \gU_h$ for the model. 

For every hard node $v \in \gU_h$, we construct its 1-hop egonet. We consider edges that \emph{direct towards} $v$ from other hard nodes $v' \in \gU_h$. This ensures that representative examples $x_i$, that are likely to be retrieved during ICL inference, have denser egonets. We experiment with both 1-hop and 2-hop set constructions. In the latter case, each node $v$ is represented by its egonet along with union of the egonets of its neighbors. 
One hyperparameter that controls the quality of the generated egonets is $m$, which is used during the construction of graph $\gG_m$. In order to determine $m$, we employ a heuristic rule based on the desired maximum iterations $\hat{T}$ until the budget $B$ is exhausted, as well as the minimum number of hard examples $N_{\hat{\theta}}$ to be covered at each iteration. Due to space limitations, additional analysis of our approach are provided in Appendices~\ref{app:Adaicl-A} and~\ref{app:Abla-D}. 

\textbf{Greedy Optimization.}
The \mcp problem is expressed as 
\begin{align}
    \text{maximize} \; &  \sum_{x_i \in \gU_h} c_i, \label{eq:mcp-1}
    \end{align}
\begin{align}
    \text{where} \; & c_j \in \{0,1\},  s_i \in \{0,1\}, \label{eq:mcp-2}
    \end{align}
\begin{align}
    \text{subject to} & \; \sum s_i \leq B, \sum_{x_j \in S_i} s_i \geq c_j. \label{eq:mcp-3}
\end{align}
\Eqref{eq:mcp-1} maximizes the coverage of the hard examples $\gU_h$, indicator variable $c_j$ denotes if example $x_j$ is covered, and $s_i$ denotes if set $S_i$ is selected. The goal is to select the examples that convey new information to the model (measured by the indicator $c_j$).
\Eqref{eq:mcp-3} ensures that we select at most $B$ sets and covered examples belong to at least one selected set (the hard examples covered in more sets are selected before others). 
To adjust the problem in our scenario, selecting set $S_i$, i.e., \mcp marks $s_i = 1$, means that we select example $x_i$ for the annotated set $\gL$. 

The \mcp problem is known to be NP-hard \citep{vazirani2001approximation}. A natural greedy solution for the \mcp chooses sets according to one rule: at each stage, choose a set that contains the largest number of uncovered elements. This approximation algorithm is summarized below in \Algref{alg:greedy}, and is well-known to approximately solve \mcp and can be further improved due to its submodularity~\citep{krause2005note} .
\begin{wrapfigure}{R}{0.62\textwidth}
\vspace{-0.3in}
    \begin{minipage}{0.62\textwidth}
     \input{algs/greedy}
    \end{minipage}
\vspace{-0.1in}
 \end{wrapfigure}

Note that the greedy algorithm is terminated when every hard example is covered, regardless of whether the budget $B$ is exhausted. In this case, the selected examples are added to the annotation set $\gL$, and the model's feedback is re-evaluated to define the new hard set $\gU'_h$. The iterative process is terminated when the total budget $B$ is exhausted. The overall framework is summarized in \Figref{fig:framework}, and the algorithm is summarized in Appendix~\ref{app:algs}.

\subsection{\adaiclp: Dynamically Re-Weighted \mcp}
The greedy \Algref{alg:greedy} for \adaicl may cover all hard examples if the budget allows. However, this might include selecting sets that contain very few hard examples, e.g., outliers, or sets that belong to isolated sparse regions. 

\adaiclp tackles this pitfall by a re-weighting schema for the \mcp problem. Whenever a hard example is covered, instead of being marked as covered, \adaiclp reduces its weight. By having new weights, dense regions with hard examples are preferred over sparse regions if their total weight is greater. We provide such a case in Appendix~\ref{app:algs}.

Unfortunately, \emph{dynamically} updating the weights of each example does not satisfy the submodularity property of \mcp, which is satisfied for \emph{fixed} weights. Nevertheless, such that we can use the greedy algorithm to approximate the optimal solution, we propose a re-weighting trick by reusing $\gU_h$ multiple times. Specifically, we copy the set $\gU_h$ multiple times, i.e., to $\gU^0_h, \gU^1_h$, etc., where different sets have different weights for their elements. If hard example $x_i$ is covered in $\gU^0_h$, then we use its weights from the other sets. Formally, we optimize 
\begin{align}
    \argmax  \sum_{t=0}^{\lfloor B/T \rfloor}\sum_{x^{t}_i \in U^{t}_h} w^t c^t_i,
    \label{eq:wmcp}
\end{align}
We set the weights $w^t$, so that $w^t \approx w^t + w^{t+1} + w^{t+2} + \cdots$, which can be achieved by exponentially reducing the weights. In our case, we set $w^t = 10^{-t}$. In the beginning, every hard example of $\gU^0_h$ has weight $w^0 = 1$ . If one example is covered in $\gU^0_h$, i.e., $c^0_i = 1$, then its new weight is obtained from $\gU^1_h$, where $w^1 = 0.1$. We introduce a new hyperparameter $T$, which denotes the desired total number of iterations until we exhaust the budget. At each iteration, we can annotate $\lfloor B/T \rfloor$ new examples by solving \Eqref{eq:wmcp} and then, the model $M$ re-evaluates its predictions. \adaiclp algorithm is summarized in Appendix~\ref{app:algs}.

\section{Experimental Setting}
With our experimental analysis, we address the following research questions (RQs):
\begin{itemize}[leftmargin=*]
    \item {RQ1}. How does \adaicl compare with SOTA active learning strategies for ICL?
    \item {RQ2}. How efficient is \adaicl regarding labeling and inference costs? 
    \item {RQ3}. How robust is \adaicl under different setups of the problem (\Figref{fig:setting})? 
    \item {RQ4}. Does \adaicl help the LLM understand the task? 
\end{itemize}

\textbf{Datasets.}
We performed empirical evaluation  with nine NLP datasets that cover well-studied tasks, such as topic classification (AGNews~\citep{zhang2015agnews}, TREC~\citep{hovy2001trec}), sentiment analysis (SST2~\citep{socher2013sst2}, Amazon~\citep{mcauley2013amazon}), natural language inference (RTE~\citep{bentivogli2009rte}, MRPC~\citep{dolan2004mrpc}, MNLI~\citep{williams2018mnli}), text summarization (XSUM~\citep{narayan2018xsum}) and math reasoning (GSM8K~\citep{cobbe2021gsm8k}). We provide examples of these datasets and additional details in Appendix~\ref{app:Setting-B}. 

\textbf{Baselines.}
We use the following approaches as baselines for comparison: (i) \textbf{Random} performs random example selection for annotation. (ii) \textbf{Pseudo-labeling} uses the LLM to generate pseudo-labels for the unlabeled instances as additional annotated data. (iii) \textbf{Fast-vote$k$} \citep{su2022selective} is a diversity-based sampling strategy that selects representative examples in the similarity space. (iv) \textbf{Vote$k$} \citep{su2022selective} additionally  accounts for the model's feedback. It sorts the examples based on the model's confidence scores and  stratifies them into $B$ equally-sized buckets. It selects the top-scoring fast-vote$k$ example from each bucket.  (v) \textbf{Hardest} resembles the uncertainty sampling strategy of active learning. The examples that the model is the most uncertain for are selected. Additionally, we include (vi) \textbf{\adaiclb} method (\Secref{sec:base}) as further ablations.

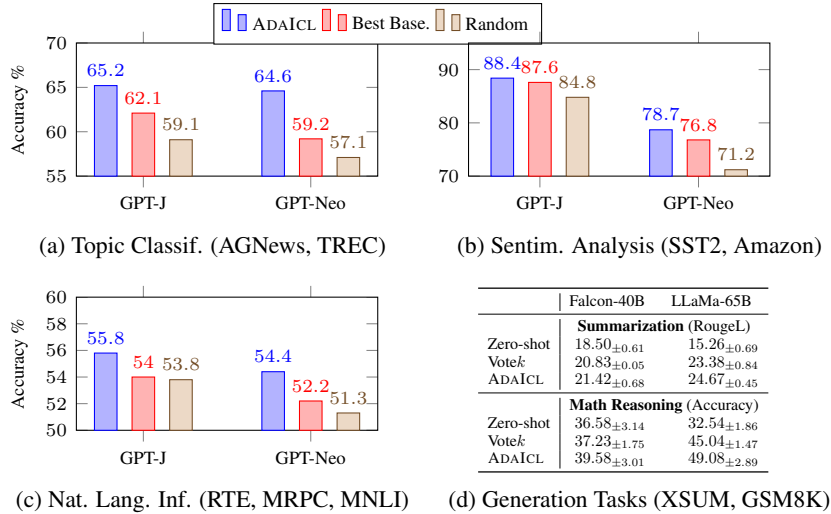
\begin{figure}
  \centering
  \begin{subfigure}[b]{0.4\textwidth}
  \pgfplotstableread[row sep=\\,col sep=&]{
Dataset & Ada  & BB & Ran \\
GPT-J    & 65.2  & 62.1  & 59.1 \\
GPT-Neo   & 64.6  & 59.2  & 57.1 \\
}\tcdata

\pgfplotstableread[row sep=\\,col sep=&]{
Dataset & Ada  & BB & Ran \\
GPT-J    & 88.4 & 87.6  & 84.8 \\
GPT-Neo    & 78.7 & 76.8 & 71.2 \\
}\sadata

\pgfplotstableread[row sep=\\,col sep=&]{
Dataset & Ada  & BB & Ran \\
GPT-J & 55.8 & 54.0 & 53.8 \\
GPT-Neo & 54.4 & 52.2 & 51.3 \\
}\nlidata

\pgfplotstableread[row sep=\\,col sep=&]{
Dataset & Ada  & BB & Ran \\
GPT-J   & 69.8 & 67.9 & 65.9 \\
GPT-Neo   & 65.9 & 62.7 & 59.9 \\
}\alldata

\begin{tikzpicture}
\tikzstyle{every node}=[font=\scriptsize]
    \begin{axis}[
            ybar=.2cm,
            bar width=.3cm,
            enlarge x limits={0.4},
            width=\textwidth,
            height=0.6\textwidth,
            legend style={at={(1,1.3)},
                anchor=north,legend columns=-1},
            symbolic x coords={ GPT-J, GPT-Neo},
            xtick=data,
            nodes near coords,
            nodes near coords align={vertical},
            ymin=55,ymax=70,
            ylabel={Accuracy \%},
            ylabel near ticks,
            compat=1.5, 
        ]
	\addplot table[x=Dataset,y=Ada]{\tcdata};
        \addplot table[x=Dataset,y=BB]{\tcdata};
        \addplot table[x=Dataset,y=Ran]{\tcdata};
        \legend{\adaicl, Best Base., Random}
    \end{axis}

\end{tikzpicture}
    \caption{Topic Classif. (AGNews, TREC)}
     \label{fig:maina}
    \vspace{0.1in}
  \end{subfigure}
  \begin{subfigure}[b]{0.4\textwidth}
  \pgfplotstableread[row sep=\\,col sep=&]{
Dataset & Ada  & BB & Ran \\
GPT-J    & 65.2  & 61.7  & 59.1 \\
GPT-Neo   & 64.6  & 59.2  & 57.1 \\
}\tcdata

\pgfplotstableread[row sep=\\,col sep=&]{
Dataset & Ada  & BB & Ran \\
GPT-J    & 88.4 & 87.6  & 84.8 \\
GPT-Neo    & 78.7 & 76.8 & 71.2 \\
}\sadata

\pgfplotstableread[row sep=\\,col sep=&]{
Dataset & Ada  & BB & Ran \\
GPT-J & 55.8 & 54.0 & 53.8 \\
GPT-Neo & 54.4 & 52.2 & 51.3 \\
}\nlidata

\pgfplotstableread[row sep=\\,col sep=&]{
Dataset & Ada  & BB & Ran \\
GPT-J   & 69.8 & 67.9 & 65.9 \\
GPT-Neo   & 65.9 & 62.7 & 59.9 \\
}\alldata

\begin{tikzpicture}
\tikzstyle{every node}=[font=\scriptsize]
    \begin{axis}[
            ybar=.2cm,
            bar width=.3cm,
            enlarge x limits={0.45},
            width=\textwidth,
            height=0.6\textwidth,
            legend style={at={(1,1)},
                anchor=north,legend columns=-1},
            symbolic x coords={ GPT-J, GPT-Neo},
            xtick=data,
            nodes near coords,
            nodes near coords align={vertical},
            ymin=70,ymax=95,
            compat=1.5, 
        ]
	\addplot table[x=Dataset,y=Ada]{\sadata};
        \addplot table[x=Dataset,y=BB]{\sadata};
        \addplot table[x=Dataset,y=Ran]{\sadata};
    \end{axis}

\end{tikzpicture}
  \caption{Sentim. Analysis (SST2, Amazon)}
   \label{fig:mainb}
    \vspace{0.1in}
  \end{subfigure}
  
  \begin{subfigure}[b]{0.4\textwidth}
  \pgfplotstableread[row sep=\\,col sep=&]{
Dataset & Ada  & BB & Ran \\
GPT-J    & 65.2  & 61.7  & 59.1 \\
GPT-Neo   & 64.6  & 59.2  & 57.1 \\
}\tcdata

\pgfplotstableread[row sep=\\,col sep=&]{
Dataset & Ada  & BB & Ran \\
GPT-J    & 88.4 & 87.6  & 84.8 \\
GPT-Neo    & 78.7 & 76.8 & 71.2 \\
}\sadata

\pgfplotstableread[row sep=\\,col sep=&]{
Dataset & Ada  & BB & Ran \\
GPT-J & 55.8 & 54.0 & 53.8 \\
GPT-Neo & 54.4 & 52.2 & 51.3 \\
}\nlidata

\pgfplotstableread[row sep=\\,col sep=&]{
Dataset & Ada  & BB & Ran \\
GPT-J   & 69.8 & 67.9 & 65.9 \\
GPT-Neo   & 65.9 & 62.7 & 59.9 \\
}\alldata

\begin{tikzpicture}
\tikzstyle{every node}=[font=\scriptsize]
    \begin{axis}[
            ybar=.2cm,
            bar width=.3cm,
            enlarge x limits={0.4},
            width=\textwidth,
            height=0.6\textwidth,
            legend style={at={(1,1)},
                anchor=north,legend columns=-1},
            symbolic x coords={ GPT-J, GPT-Neo},
            xtick=data,
            nodes near coords,
            nodes near coords align={vertical},
            ymin=50,ymax=60,
            ylabel={Accuracy \%},
            ylabel near ticks,
            compat=1.5, 
        ]
	\addplot table[x=Dataset,y=Ada]{\nlidata};
        \addplot table[x=Dataset,y=BB]{\nlidata};
        \addplot table[x=Dataset,y=Ran]{\nlidata};
    \end{axis}

\end{tikzpicture}
  \caption{Nat. Lang. Inf. (RTE, MRPC, MNLI)}
   \label{fig:mainc}
  \end{subfigure}
  \begin{subfigure}[b]{0.4\textwidth}
  \hspace{.2in}
  \resizebox{0.7\linewidth}{!}{
  \centering

\centering
\begin{tabular}{l|rr}
    \toprule
    & \multicolumn{1}{c}{Falcon-40B} & \multicolumn{1}{c}{LLaMa-65B} \\
    \midrule
    & \multicolumn{2}{c}{\textbf{Summarization} (RougeL)} \\
    Zero-shot & $18.50_{\pm 0.61}$ & $15.26_{\pm 0.69}$ \\
    Vote$k$ & $20.83_{\pm 0.05}$ & $23.38_{\pm 0.84}$ \\
   \adaicl & $21.42_{\pm 0.68}$ & $24.67_{\pm 0.45}$ \\
    \midrule
    & \multicolumn{2}{c}{\textbf{Math Reasoning} (Accuracy)} \\
    Zero-shot &  $36.58_{\pm 3.14}$ & $32.54_{\pm 1.86}$ \\
    Vote$k$ &  $37.23_{\pm 1.75}$ & $45.04_{\pm 1.47}$ \\ 
   \adaicl &  $39.58_{\pm 3.01}$ & $49.08_{\pm 2.89}$  \\
    \bottomrule
\end{tabular}%

}
  \caption{Generation Tasks (XSUM, GSM8K)}
   \label{fig:maind}
  \end{subfigure}
  \caption{Performance comparison across different tasks with GPT-J (6B) and GPT-Neo (1.3B). ``\textit{Best Base.}'' denotes the \underline{best} baseline for the task. \adaicl performs the best, while for the classification tasks \adaiclb is the second-best (full results in Appendix~\ref{app:full_res}).
  }
    \label{fig:main}
\end{figure} 

\textbf{Design Space.}
As summarized in \Figref{fig:setting}, the design space includes the unlabeled set $\gU$, the number of ICL examples $k$, the similarity space $\gS$, the budget $B$, and the LLM $M$.
We experiment with seven LLMs of varying sizes (1.3B to 65B parameters), including GPT, Mosaic, Falcon, and LLaMa model families,  all of which are open-source and allow the reproducibility of our research.
Unless otherwise stated, we set $k=5$, $B=20$ and we obtain embeddings in the similarity space via SBERT~\citep{reimers2019sentence}.
We experiment with inductive settings, where test instances come from an \emph{unseen} set $\gU_{\text{test}}$, but also for transductive settings, where test instances come from $\gU$.

\textbf{\adaicl}.
 For the default problem setup, we construct 2-hop sets with $m=5$ for \adaicl, and 1-hop sets with $m=15$ for \adaiclp via \Eqref{eq:heuristic}.  The default number of iterations $T$ for \adaiclp is $T=2$, while in the additive budget scenario we have $T=1$. As the threshold hyper-parameter $\theta$, we have $\theta=0.5$, i.e., 50\% of the examples are considered as hard. Hyper-parameter sensitivity studies that show \adaicl's robustness are presented in Appendix~\ref{app:Abla-D}.

\section{Results \& Analysis}

\subsection{RQ1: \adaicl is Effective} \label{sec:res1}

Figures~\ref{fig:maina},~\ref{fig:mainb}, and ~\ref{fig:mainc} show performance results for classification tasks with two different models GPT-J (6B) and GPT-Neo (1.3B). \adaicl is the method that achieves the best performance, with an improvement of up to 7.5\%  accuracy points over random selection. The overall improvement over the best baseline is 1.9\% points for GPT-J and 3.2\% for GPT-Neo, which shows that \adaicl is important for smaller sized LMs. 
The second best performing method for topic classification and sentiment analysis is \adaiclb, which shows the  importance of diversity-based uncertainty sampling for ICL.
Figure~\ref{fig:maind} provides results for generation tasks. On the challenging reasoning tasks, \adaicl outperforms Vote$k$ and zero-shot ICL by 4.04\% and 16.54\% in accuracy, respectively.

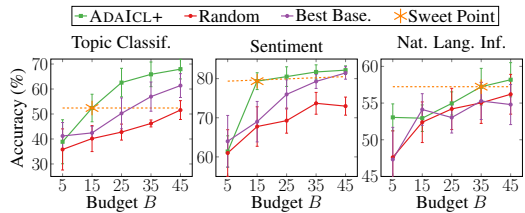
\begin{wrapfigure}{r}{0.5\textwidth}
  \centering
    \resizebox{0.5\columnwidth}{!}{\definecolor{col1}{rgb}{0.60, 0.31, 0.64}
\definecolor{col2}{rgb}{0.30, 0.69, 0.29}
\definecolor{col3}{rgb}{0.22, 0.49, 0.72}
\definecolor{col4}{rgb}{0.89, 0.10, 0.11}
\definecolor{col5}{rgb}{1, 1, 0.8}

\begin{tikzpicture}
\tikzstyle{every node}=[font=\Huge]

\begin{axis}[legend style={at={(.9,0.9),anchor=north east}},
             legend style={legend pos=outer north east,},  title={Topic Classif.},ylabel={Accuracy (\%)}, xlabel={Budget $B$ }, every axis plot/.append style={ultra thick}, ymin=25, ymax=72, compat=1.5, xtick={5,15,25,35,45}
]

\addplot[mark=square*,col2, error bars/.cd, y dir=both, y explicit] coordinates {
    (5, 38.89) +- (5, 8.81) 
    (15, 52.40) +- (15, 5.51)
    (25, 62.57) +- (25, 5.72)
    (35, 65.88) +- (35, 4.94)
    (45, 67.95) +- (45, 3.91)
};

\addplot[dashed, orange] coordinates {(5, 52.40)(45,52.40)};

\addplot[mark=*,col4, error bars/.cd, y dir=both, y explicit] coordinates {
    (5, 35.81) +- (5, 8.23)
    (15, 40.16) +- (15, 5.20)
    (25, 42.76) +- (25, 3.20)
    (35, 46.22) +- (35, 1.39)
    (45, 51.58) +- (45, 3.77)
};

\addplot[mark=*,col1, error bars/.cd, y dir=both, y explicit] coordinates {
    (5, 41.2) +- (5, 5.40)
    (15, 42.44) +- (15, 2.61)
    (25, 50.25) +- (25, 6.19)
    (35, 56.96) +- (35, 5.85)
    (45, 61.38) +- (45, 4.77)
};

\addplot[only marks, mark=asterisk,orange, mark size=10] 
coordinates {
(15, 52.40)
};
\end{axis}

\begin{axis}[legend columns=-1,legend style={at={(2,1.45),anchor=north east,}},  title={Sentiment}, xlabel={Budget $B$ }, ymin=55, ymax=85,  every axis plot/.append style={ultra thick},  xshift=8cm, compat=1.5, xtick={5,15,25,35,45}
]
\addplot[mark=square*,col2, error bars/.cd, y dir=both, y explicit] coordinates {
    (5, 61.38) +- (5, 4.01) 
    (15, 79.35) +- (15, 2.15)
    (25, 80.51) +- (25, 2.51)
    (35, 81.70) +- (35, 1.82)
    (45, 82.16) +- (45, 1.10)
};
\addlegendentry{\adaiclp \; \; }

\addplot[mark=*,col4, error bars/.cd, y dir=both, y explicit] coordinates {
    (5, 60.99) +- (5, 5.96)
    (15, 67.76) +- (15, 5.05)
    (25, 69.26) +- (25, 3.20)
    (35, 73.69) +- (35, 2.79)
    (45, 72.97) +- (45, 2.31)
};
\addlegendentry{Random \; \; } 

\addplot[mark=*,col1, error bars/.cd, y dir=both, y explicit] coordinates {
    (5, 63.99) +- (5, 6.60)
    (15, 69.00) +- (15, 5.17)
    (25, 75.97) +- (25, 3.72)
    (35, 79.28) +- (35, 1.79)
    (45, 81.44) +- (45, 1.59)
};
\addlegendentry{Best Base. \; \; }

\addplot[only marks, mark=asterisk,orange, mark size=10] 
coordinates {
(15, 79.35)
};
\addlegendentry{Sweet Point}

\addplot[dashed, orange] coordinates {(5, 79.35)(45,80.51)};

\end{axis}

\begin{axis}[legend style={at={(.9,0.9),anchor=north east}},
             legend style={legend pos=outer north east,},  title={Nat. Lang. Inf.}, xlabel={Budget $B$ }, every axis plot/.append style={ultra thick}, ymin=45, ymax=61, xshift=16cm, xtick={5,15,25,35,45}, compat=1.5, 
]
\addplot[mark=square*,col2, error bars/.cd, y dir=both, y explicit] coordinates {
    (5, 53.05) +- (5, 1.84) 
    (15, 52.94) +- (15, 1.81)
    (25, 54.95) +- (25, 1.58)
    (35, 57.23) +- (35, 2.50)
    (45, 58.18) +- (45, 2.33)
};

\addplot[dashed, orange] coordinates {(5, 57.23)(45,57.23)};

\addplot[mark=*,col4, error bars/.cd, y dir=both, y explicit] coordinates {
    (5, 47.61) +- (5, 3.61)
    (15, 52.39) +- (15, 2.77)
    (25, 54.20) +- (25, 2.81)
    (35, 55.03) +- (35, 2.80)
    (45, 56.20) +- (45, 2.71)
};

\addplot[mark=*,col1, error bars/.cd, y dir=both, y explicit] coordinates {
    (5, 47.32) +- (5, 4.37)
    (15, 54.12) +- (15, 2.16)
    (25, 53.03) +- (25, 2.11)
    (35, 55.28) +- (35, 2.52)
    (45, 54.80) +- (45, 2.73)
};
\addplot[only marks, mark=asterisk,orange, mark size=10] 
coordinates {
(35, 57.23)
};
\end{axis}

\end{tikzpicture}}
      \caption{Multi-step results with GPT-Neo. Sweet point: the point at which we exceed the best performance achieved by random selection. 
      }
      \label{fig:phases}
    \vspace{-0.17in}
\end{wrapfigure}

\subsection{RQ2 \& RQ3: \adaicl is Efficient and Robust} \label{sec:res2}

\textbf{Budget-Efficiency}. We experiment with a scenario similar to mini-batch active learning, where the budget increases in different steps
and the retriever uses as many ICL annotated examples as the context-length limit allows. \Figref{fig:phases} shows results when incrementing the budget with 10 more annotations (for 4 steps). \adaicl performs the best in all cases, where the average accuracy improvement over the best baseline is 7.09\% for topic classification, 3.08\% for sentiment analysis, and 2.36\% for natural language inference. \Figref{fig:phases}  also shows that for topic classification and sentiment analysis \adaicl exceeds the best performance achieved by random annotation with $3\times$ less budget.

\begin{wraptable}{r}{0.5 \textwidth} %
\centering
\caption{Impact of the number of ICL examples.}
\label{tab:kshot}%
\resizebox{\linewidth}{!}{
\begin{threeparttable}
    \vspace{-0.17in}
    \begin{tabular}{l|cc|cc}
        \toprule
         & \multicolumn{2}{c|}{GPT-J (6B)} & \multicolumn{2}{c}{Mosaic (7B)}  \\
        & AGNews & SST2 & AGNews & SST2 \\
        \midrule
        Vote$k$ 5-shot & $53.61_{\pm 7.72}$ & $72.89_{\pm 11.67}$ & $71.04_{\pm 8.12}$  & $	80.98_{\pm 5.26}$ \\
        Vote$k$ 10-shot & $58.32_{\pm 2.74}$ & $80.08_{\pm 7.43}$& $76.09_{\pm 3.37}$ & $90.23_{\pm 0.31}$\\
        \adaicl 5-shot & $67.44_{\pm 4.57}$& $83.98_{\pm 1.10}$ & $77.20_{\pm 3.42}$& $89.58_{\pm 1.75}$\\
        
        \bottomrule
    \end{tabular}%
		\begin{tablenotes}
  \item $B=5$ and $B=10$ for 5-shot and 10-shot ICL, respectively.
        \end{tablenotes}
    \end{threeparttable}
}

    \vspace{-0.1in}
\end{wraptable}%

\textbf{Impact of ICL examples}. Table~\ref{tab:kshot} investigates \adaicl's efficiency with respect to the number of ICL examples used during inference (and annotated). As shown, \adaicl outperforms vote$k$ although it uses and annotates $2 \times$ fewer ICL examples. This indicates that \adaicl identifies examples that help the LLM learn the task, while it can reduce the inference costs due to shorter input prompts.  We provide a visualization of \adaicl's process in Appendix~\ref{app:visual}.

\textbf{Retriever Effect}. Table~\ref{tab:sbert} shows results when we use different off-the-shelf encoders for the similarity space $\gS$, such as BERT~\citep{devlin2019bert} and RoBERTa~\citep{liu2019roberta}. As discussed in \Secref{sec:challenges}, it is difficult to approximate the true similarity between examples based on the pretraining distribution $\gD$, and thus different encoders lead to different results. 
For example, SBERT achieves a maximum average performance of 55.33\% and 77.73\% for TREC and Amazon, respectively, while BERT achieves 44.06\% and 85.80\%. Despite the encoder choice, \adaicl performs overall the best as its diversity-based uncertainty sampling mitigates this effect.

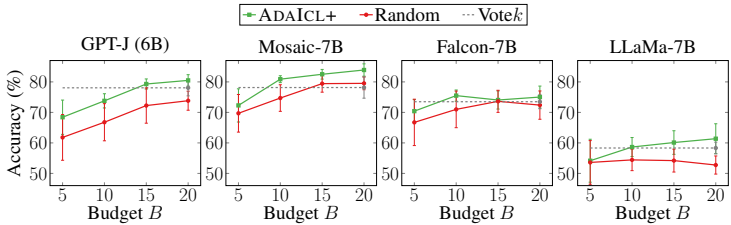
\begin{wrapfigure}{r}{0.7\textwidth}
  \centering
    \resizebox{0.7\textwidth}{!}{\definecolor{col1}{rgb}{0.60, 0.31, 0.64}
\definecolor{col2}{rgb}{0.30, 0.69, 0.29}
\definecolor{col3}{rgb}{0.22, 0.49, 0.72}
\definecolor{col4}{rgb}{0.89, 0.10, 0.11}
\definecolor{col5}{rgb}{1, 1, 0.8}

\begin{tikzpicture}
\tikzstyle{every node}=[font=\Huge]

\begin{axis}[legend style={at={(.9,0.9),anchor=north east}},
             legend style={legend pos=outer north east,},  title={GPT-J (6B)},ylabel={Accuracy (\%)}, xlabel={Budget $B$ }, every axis plot/.append style={ultra thick}, ymin=46, ymax=87, compat=1.5, xtick={5,10,15,20}
]

\addplot[mark=square*,col2, error bars/.cd, y dir=both, y explicit] coordinates {
    (5, 68.42) +- (5, 5.62)
    (10, 73.79) +- (10, 2.33)
    (15, 79.31)  +- (15, 1.66)
    (20, 80.49) +- (20, 1.88)
};

\addplot[dashed, gray] coordinates {(5, 78.06)(20,78.06)};

\addplot[mark=*,col4, error bars/.cd, y dir=both, y explicit] coordinates {
    (5, 61.80) +- (5, 7.49)
    (10, 66.78) +- (10, 6.09)
    (15, 72.24) +- (15, 5.78)
    (20, 73.83) +- (20, 3.12)
};

\addplot[only marks, mark=*,gray, error bars/.cd, y dir=both, y explicit] 
coordinates {
(20, 78.06) +- (20, 2.64)
};

\end{axis}

\begin{axis}[legend columns=-1,legend style={at={(2,1.45),anchor=north east,}},  title={Mosaic-7B}, xlabel={Budget $B$ }, ymin=46, ymax=87,  every axis plot/.append style={ultra thick},  xshift=8cm, compat=1.5, xtick={5,10,15,20}
]
\addplot[mark=square*,col2, error bars/.cd, y dir=both, y explicit] coordinates {
    (5, 72.29) +- (5, 5.44)
    (10, 80.89) +- (10, 1.13)
    (15, 82.50) +- (15, 1.59)
    (20, 83.90) +- (20, 2.02)
    
};
\addlegendentry{\adaiclp \; \; }

\addplot[mark=*,col4, error bars/.cd, y dir=both, y explicit] coordinates {
    (5, 69.71) +- (5, 6.17)
    (10, 74.71) +- (10, 4.36)
    (15, 79.43) +- (15, 2.80)
    (20, 79.52) +- (20, 2.09)
};
\addlegendentry{Random \; \; }

\addplot[dashed, gray] coordinates {(5, 78.15)(20,78.15)};
\addlegendentry{Vote$k$} 

\addplot[only marks, mark=*,gray, error bars/.cd, y dir=both, y explicit] 
coordinates {
(20, 78.15) +- (20, 3.49)
};

\end{axis}

\begin{axis}[legend style={at={(.9,0.9),anchor=north east}},
             legend style={legend pos=outer north east,},  title={Falcon-7B}, xlabel={Budget $B$ }, every axis plot/.append style={ultra thick}, ymin=46, ymax=87, xshift=16cm, xtick={5,10,15,20}, compat=1.5, 
]
\addplot[mark=square*,col2, error bars/.cd, y dir=both, y explicit] coordinates {
    (5, 70.41) +- (5, 3.77)
    (10, 75.50) +- (10, 1.85)
    (15, 74.06) +- (15, 2.97)
    (20, 75.03) +- (20, 3.60)
};

\addplot[mark=*,col4, error bars/.cd, y dir=both, y explicit] coordinates {
    (5, 66.73) +- (5, 7.60)
    (10, 70.99) +- (10, 6.00)
    (15, 73.63) +- (15, 3.62)
    (20, 72.39) +- (20, 4.64)
    
};

\addplot[only marks, mark=*,gray, error bars/.cd, y dir=both, y explicit] 
coordinates {
(20, 73.50) +- (20, 2.29)
};
\addplot[dashed, gray] coordinates {(5, 73.50)(20, 73.50)};

\end{axis}

\begin{axis}[legend style={at={(.9,0.9),anchor=north east}},
             legend style={legend pos=outer north east,},  title={LLaMa-7B}, xlabel={Budget $B$ }, every axis plot/.append style={ultra thick}, ymin=46, ymax=87, xshift=24cm, xtick={5,10,15,20}, compat=1.5, 
]
\addplot[mark=square*,col2, error bars/.cd, y dir=both, y explicit] coordinates {

    (5, 54.16) +- (5, 7.04)
    (10, 58.65) +- (10, 3.12)
    (15, 60.12) +- (15, 3.87)
    (20, 61.38) +- (20, 4.90)
};

\addplot[mark=*,col4, error bars/.cd, y dir=both, y explicit] coordinates {
    (5, 53.60) +- (5, 7.16)
    (10, 54.43) +- (10, 3.52)
    (15, 54.20) +- (15, 3.75)
    (20, 52.75) +- (20, 2.97)
};

\addplot[only marks, mark=*,gray, error bars/.cd, y dir=both, y explicit] 
coordinates {
(20, 58.33) +- (20, 1.91)
};
\addplot[dashed, gray] coordinates {(5, 58.33)(20, 58.33)};

\end{axis}

\end{tikzpicture}}
    \vspace{-0.17in}
     \caption{Average results over AGNews, TREC, SST2, and Amazon datasets for four LLMs with similar size.}
     \label{fig:llms}
     \vspace{-0.17in}
\end{wrapfigure}
\textbf{LLM Effect}. \Figref{fig:llms} shows results when using different LLMs of similar sizes (6-7B parameters). The best performance is achieved for the Mosaic and GPT-J models. LLaMa does not display effective ICL capabilities and Falcon does not substantially improve with more annotated examples. For Mosaic and GPT-J models, \adaicl outperforms Vote$k$ by 4.09\% accuracy points, while the average accuracy improvement over random annotation for different budgets is 5.45\%  points.

\begin{table}[tb]
\centering

\caption{Performance comparison across different retrieval and semantic similarity configurations.}
\label{tab:sbert}%
\resizebox{\linewidth}{!}{
\begin{threeparttable}
\vspace{-0.17in}
    \begin{tabular}{l|ccc|ccc|ccc|c}
        \toprule
         Retriever, $\gS \xrightarrow{}$ & \multicolumn{3}{c|}{SBERT-all-mpnet-base-v2} & \multicolumn{3}{c|}{RoBERTa-nli-large-mean-tokens} & \multicolumn{3}{c|}{BERT-nli-large-cls-pool} & Avg. \\
        & TREC & SST2 & Amazon & TREC & SST2 & Amazon & TREC & SST2 & Amazon &  \\
        \midrule
        Pseudo-labeling & $48.56_{\pm 6.33}$& $69.13_{\pm 3.87}$ & $70.96_{\pm 3.35}$ & $33.98_{\pm 3.68}$ & $74.08_{\pm 4.40}$& $81.11_{\pm 4.14}$ & $41.27_{\pm 4.24}$ & $77.47_{\pm 1.60}$ & $81.63_{\pm 2.49}$ & $64.24$\\
        Random &  $54.68_{\pm 1.68}$ & $68.48_{\pm 1.87}$ & $73.95_{\pm 2.03}$ & $37.23_{\pm 2.30}$ & $74.21_{\pm 3.50}$ & $84.46_{\pm 3.21}$ & $34.75_{\pm 2.41}$ & $72.65_{\pm 5.82}$ & $80.20_{\pm 3.34}$ & $64.51$\\
        Vote$k$ & $54.81_{\pm 0.49}$ & $73.69_{\pm 9.05}$ & $ 75.13_{\pm 0.98}$ & $37.77_{\pm 4.65}$ &$76.16_{\pm 2.23}$ & $84.11_{\pm 1.28}$& $42.43_{\pm 3.34}$ & $80.85_{\pm 2.09}$& $83.59_{\pm 1.77}$ & $67.61$\\
        \adaiclb &  $48.24_{\pm 0.98}$ & $77.86_{\pm 1.02}$ & $75.77_{\pm 3.63}$ & $38.12_{\pm 5.74}$ & $78.12_{\pm 5.30}$ & $\underline{85.93}_{\pm 2.30}$ & $	38.15_{\pm 3.10}$ & $78.64_{\pm 2.78}$ & $\underline{85.80}_{\pm 1.75}$ & $67.40$\\ 
       \textbf{\adaicl} &    $\underline{55.33}_{\pm 2.57}$ & $\underline{79.68}_{\pm 2.47}$ & $\underline{77.73}_{\pm 2.23}$ & $\underline{39.06}_{\pm 3.37}$ & $\underline{81.11}_{\pm 1.50}$ & $ 85.15_{\pm 0.55}$ & $\underline{44.06}_{\pm 2.49}$ & $\underline{80.85}_{\pm 2.83}$ & $84.65_{\pm 3.52}$ & $\textbf{69.74}$\\
        \bottomrule
    \end{tabular}%
    \end{threeparttable}
}
\vspace{-0.17in}

\end{table}%

\subsection{RQ3: \adaicl improves Calibration} \label{sec:res-calib}

One motivation behind uncertainty sampling is that it helps the LLM understand the task. Our hypothesis is that selecting examples that the LLM is over-confident for, i.e., easy examples, might not convey new information for the LLM to truly understand the task. We test our hypothesis by introducing a new variant of our method, termed \adaicl-easy, which constructs regions of hard examples \emph{around easy examples} for \mcp. To compare \adaicl-easy with our original \adaicl-hard, we compute the expected calibration error (ECE)~\citep{guo2017calibration} which quantifies the discrepancy between a model's predicted probabilities (how well it believes it understands the task) and observed outcomes (how well it actually solves the task). In addition, we visualize the discrepancy across probability bins by reliability plots, where any deviation from the straight diagonal indicates miscalibration and misunderstanding of the task.

\begin{figure*}[h]
    \centering
    \includegraphics[width=\linewidth]{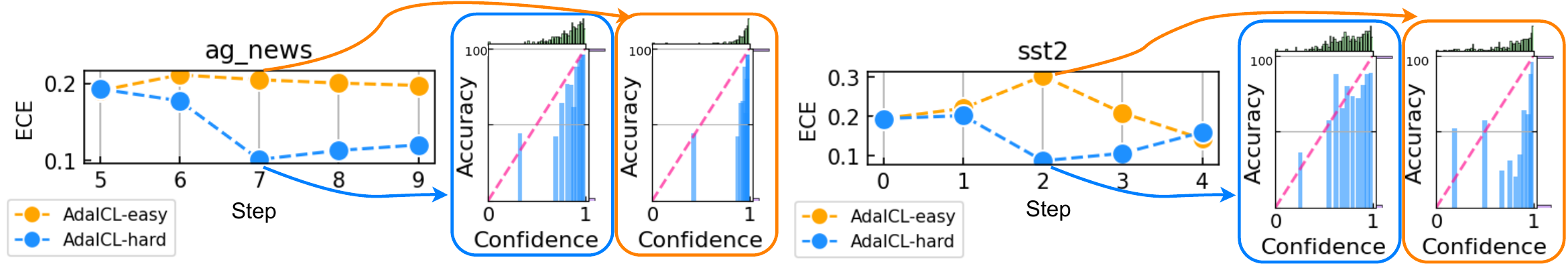}
    \caption{Expected calibration errors (ECE -- the lower the better) and reliability plots for \adaicl when selecting easy (\adaicl-easy) and  selecting hard (\adaicl-hard) examples.}
    \label{fig:calibr}
\end{figure*}

As shown in \Figref{fig:calibr}, ``phase changes'' in ECE could be observed for \adaicl-hard at different iterations and thus, improved calibration. We conjecture that the over-confident examples selected using \adaicl-easy lead to the bias towards making over-confident ICL predictions that are not always true. This is verified by the reliability plots for two snapshots: at iteration=7 for AGNews and at iteration=2 for SST2, where \adaicl-hard reduces the ECE from 0.20-0.30 to approximately 0.1.
\adaicl-easy tends to make over-confident predictions (heavily skewed towards the right of the x-axis and large deviation from the diagonal), whereas \adaicl-hard produces more calibrated predictions, with more uniform probability distributions. We provide additional calibration analysis in Appendix~\ref{app:calibr2}.

\section{Conclusions}
In this work, we investigated budgeted example selection for annotation for ICL, a previously under-explored area. Seeking for an effective and efficient selection, we introduce \adaicl, a diversity-based uncertainty selection strategy. Diversity-based sampling improves effectiveness, while uncertainty sampling improves budget efficiency and helps the LLM learn new information about the task. 
Extensive experiments in low-resource settings show that \adaicl outperforms other approaches in five NLP tasks using seven different LLMs. Moreover, \adaicl can result into considerable budget savings, while it also needs fewer ICL examples during inference to achieve a given level of performance, reducing inference costs. Finally, our calibration analysis showed that \adaicl selects examples that lead to well-calibrated predictions.

\section{Reproducibility Statement}
The code of our \adaicl algorithm is at \url{https://github.com/amazon-science/adaptive-in-context-learning}. \adaicl's algorithmic steps are extensively summarized in Appendix~\ref{app:Adaicl-A}. For the datasets used, a complete description of the data processing steps is given in Appendices~\ref{app:data} and~\ref{app:prompt-data}. Details of the experimental configurations are given in Appendix~\ref{app:conf}.

\bibliography{ICL}

\begin{thebibliography}{77}
\providecommand{\natexlab}[1]{#1}
\providecommand{\url}[1]{\texttt{#1}}
\expandafter\ifx\csname urlstyle\endcsname\relax
  \providecommand{\doi}[1]{doi: #1}\else
  \providecommand{\doi}{doi: \begingroup \urlstyle{rm}\Url}\fi

\bibitem[Aky{\"u}rek et~al.(2023)Aky{\"u}rek, Schuurmans, Andreas, Ma, and
  Zhou]{akyurek2022learning}
Ekin Aky{\"u}rek, Dale Schuurmans, Jacob Andreas, Tengyu Ma, and Denny Zhou.
\newblock What learning algorithm is in-context learning? investigations with
  linear models.
\newblock \emph{ICLR}, 2023.

\bibitem[Bai et~al.(2023)Bai, Chen, Wang, Xiong, and Mei]{bai2023statisticians}
Yu~Bai, Fan Chen, Huan Wang, Caiming Xiong, and Song Mei.
\newblock Transformers as statisticians: Provable in-context learning with
  in-context algorithm selection.
\newblock \emph{arXiv preprint arXiv:2306.04637}, 2023.

\bibitem[Bansal et~al.(2023)Bansal, Gopalakrishnan, Dingliwal, Bodapati,
  Kirchhoff, and Roth]{bansal2022rethinking}
Hritik Bansal, Karthik Gopalakrishnan, Saket Dingliwal, Sravan Bodapati, Katrin
  Kirchhoff, and Dan Roth.
\newblock Rethinking the role of scale for in-context learning: An
  interpretability-based case study at 66 billion scale.
\newblock \emph{ACL}, 2023.

\bibitem[Bentivogli et~al.(2009)Bentivogli, Clark, Dagan, and
  Giampiccolo]{bentivogli2009rte}
Luisa Bentivogli, Peter Clark, Ido Dagan, and Danilo Giampiccolo.
\newblock The fifth pascal recognizing textual entailment challenge.
\newblock \emph{TAC}, 2009.

\bibitem[Bommasani et~al.(2021)Bommasani, Hudson, Adeli, Altman, Arora, von
  Arx, Bernstein, Bohg, Bosselut, Brunskill,
  et~al.]{bommasani2021opportunities}
Rishi Bommasani, Drew~A Hudson, Ehsan Adeli, Russ Altman, Simran Arora, Sydney
  von Arx, Michael~S Bernstein, Jeannette Bohg, Antoine Bosselut, Emma
  Brunskill, et~al.
\newblock On the opportunities and risks of foundation models.
\newblock \emph{arXiv preprint arXiv:2108.07258}, 2021.

\bibitem[Brown et~al.(2020)Brown, Mann, Ryder, Subbiah, Kaplan, Dhariwal,
  Neelakantan, Shyam, Sastry, Askell, et~al.]{brown2020fewshot}
Tom Brown, Benjamin Mann, Nick Ryder, Melanie Subbiah, Jared~D Kaplan, Prafulla
  Dhariwal, Arvind Neelakantan, Pranav Shyam, Girish Sastry, Amanda Askell,
  et~al.
\newblock Language models are few-shot learners.
\newblock \emph{NeurIPS}, 2020.

\bibitem[Chan et~al.(2022)Chan, Santoro, Lampinen, Wang, Singh, Richemond,
  McClelland, and Hill]{chan2022data}
Stephanie Chan, Adam Santoro, Andrew Lampinen, Jane Wang, Aaditya Singh, Pierre
  Richemond, James McClelland, and Felix Hill.
\newblock Data distributional properties drive emergent in-context learning in
  transformers.
\newblock \emph{NeurIPS}, 2022.

\bibitem[Chen et~al.(2022{\natexlab{a}})Chen, Zhao, Yu, McKeown, and
  He]{chen2022relation}
Yanda Chen, Chen Zhao, Zhou Yu, Kathleen McKeown, and He~He.
\newblock On the relation between sensitivity and accuracy in in-context
  learning.
\newblock \emph{arXiv preprint arXiv:2209.07661}, 2022{\natexlab{a}}.

\bibitem[Chen et~al.(2022{\natexlab{b}})Chen, Zhong, Zha, Karypis, and
  He]{chen2021meta}
Yanda Chen, Ruiqi Zhong, Sheng Zha, George Karypis, and He~He.
\newblock Meta-learning via language model in-context tuning.
\newblock \emph{ACL}, 2022{\natexlab{b}}.

\bibitem[Cobbe et~al.(2021)Cobbe, Kosaraju, Bavarian, Chen, Jun, Kaiser,
  Plappert, Tworek, Hilton, Nakano, et~al.]{cobbe2021gsm8k}
Karl Cobbe, Vineet Kosaraju, Mohammad Bavarian, Mark Chen, Heewoo Jun, Lukasz
  Kaiser, Matthias Plappert, Jerry Tworek, Jacob Hilton, Reiichiro Nakano,
  et~al.
\newblock Training verifiers to solve math word problems.
\newblock \emph{arXiv preprint arXiv:2110.14168}, 2021.

\bibitem[Devlin et~al.(2019)Devlin, Chang, Lee, and Toutanova]{devlin2019bert}
Jacob Devlin, Ming-Wei Chang, Kenton Lee, and Kristina Toutanova.
\newblock Bert: Pre-training of deep bidirectional transformers for language
  understanding.
\newblock In \emph{NAACL}, 2019.

\bibitem[Diao et~al.(2023)Diao, Wang, Lin, and Zhang]{diao2023active}
Shizhe Diao, Pengcheng Wang, Yong Lin, and Tong Zhang.
\newblock Active prompting with chain-of-thought for large language models.
\newblock \emph{arXiv preprint arXiv:2302.12246}, 2023.

\bibitem[Dolan et~al.(2004)Dolan, Quirk, and Brockett]{dolan2004mrpc}
Bill Dolan, Chris Quirk, and Chris Brockett.
\newblock Unsupervised construction of large paraphrase corpora: Exploiting
  massively parallel news sources.
\newblock In \emph{{COLING} 2004: Proceedings of the 20th International
  Conference on Computational Linguistics}, aug 23{--}aug 27 2004.
\newblock URL \url{https://aclanthology.org/C04-1051}.

\bibitem[Dong et~al.(2022)Dong, Li, Dai, Zheng, Wu, Chang, Sun, Xu, and
  Sui]{dong2022survey-icl}
Qingxiu Dong, Lei Li, Damai Dai, Ce~Zheng, Zhiyong Wu, Baobao Chang, Xu~Sun,
  Jingjing Xu, and Zhifang Sui.
\newblock A survey for in-context learning.
\newblock \emph{arXiv preprint arXiv:2301.00234}, 2022.

\bibitem[Ein-Dor et~al.(2020)Ein-Dor, Halfon, Gera, Shnarch, Dankin, Choshen,
  Danilevsky, Aharonov, Katz, and Slonim]{ein-dor-etal-2020-active}
Liat Ein-Dor, Alon Halfon, Ariel Gera, Eyal Shnarch, Lena Dankin, Leshem
  Choshen, Marina Danilevsky, Ranit Aharonov, Yoav Katz, and Noam Slonim.
\newblock {A}ctive {L}earning for {BERT}: {A}n {E}mpirical {S}tudy.
\newblock In \emph{EMNLP}, 2020.

\bibitem[Erdmann et~al.(2019)Erdmann, Wrisley, Allen, Brown,
  Cohen-Bod{\'e}n{\`e}s, Elsner, Feng, Joseph, Joyeux-Prunel, and
  de~Marneffe]{erdmann-etal-2019-practical}
Alexander Erdmann, David~Joseph Wrisley, Benjamin Allen, Christopher Brown,
  Sophie Cohen-Bod{\'e}n{\`e}s, Micha Elsner, Yukun Feng, Brian Joseph,
  B{\'e}atrice Joyeux-Prunel, and Marie-Catherine de~Marneffe.
\newblock Practical, efficient, and customizable active learning for named
  entity recognition in the digital humanities.
\newblock In \emph{EMNLP}, 2019.

\bibitem[Gao et~al.(2021)Gao, Fisch, and Chen]{gao-etal-2021-making-prompt}
Tianyu Gao, Adam Fisch, and Danqi Chen.
\newblock Making pre-trained language models better few-shot learners.
\newblock In \emph{ACL}, 2021.

\bibitem[Garg et~al.(2022)Garg, Tsipras, Liang, and
  Valiant]{garg2022icl-linear}
Shivam Garg, Dimitris Tsipras, Percy~S Liang, and Gregory Valiant.
\newblock What can transformers learn in-context? a case study of simple
  function classes.
\newblock \emph{NeurIPS}, 2022.

\bibitem[Guo et~al.(2017)Guo, Pleiss, Sun, and Weinberger]{guo2017calibration}
Chuan Guo, Geoff Pleiss, Yu~Sun, and Kilian~Q Weinberger.
\newblock On calibration of modern neural networks.
\newblock In \emph{ICML}, 2017.

\bibitem[Haffari et~al.(2009)Haffari, Roy, and Sarkar]{haffari2009active}
Gholamreza Haffari, Maxim Roy, and Anoop Sarkar.
\newblock Active learning for statistical phrase-based machine translation.
\newblock In \emph{NAACL}, 2009.

\bibitem[Hahn \& Goyal(2023)Hahn and Goyal]{hahn2023theory}
Michael Hahn and Navin Goyal.
\newblock A theory of emergent in-context learning as implicit structure
  induction.
\newblock \emph{arXiv preprint arXiv:2303.07971}, 2023.

\bibitem[Han et~al.(2023)Han, Wang, Zhao, and Ji]{han2023icl-kernel}
Chi Han, Ziqi Wang, Han Zhao, and Heng Ji.
\newblock In-context learning of large language models explained as kernel
  regression.
\newblock \emph{arXiv preprint arXiv:2305.12766}, 2023.

\bibitem[Heese et~al.(2023)Heese, Schmid, Walczak, and
  Bortz]{heese2023calibrated}
Raoul Heese, Jochen Schmid, Micha{\l} Walczak, and Michael Bortz.
\newblock Calibrated simplex-mapping classification.
\newblock \emph{PLoS One}, 18\penalty0 (1):\penalty0 e0279876, 2023.

\bibitem[Hovy et~al.(2001)Hovy, Gerber, Hermjakob, Lin, and
  Ravichandran]{hovy2001trec}
Eduard Hovy, Laurie Gerber, Ulf Hermjakob, Chin-Yew Lin, and Deepak
  Ravichandran.
\newblock Toward semantics-based answer pinpointing.
\newblock In \emph{Proceedings of the First International Conference on Human
  Language Technology Research}, 2001.
\newblock URL \url{https://www.aclweb.org/anthology/H01-1069}.

\bibitem[Jiang et~al.(2021)Jiang, Araki, Ding, and Neubig]{jiang2021can}
Zhengbao Jiang, Jun Araki, Haibo Ding, and Graham Neubig.
\newblock How can we know when language models know? on the calibration of
  language models for question answering.
\newblock \emph{Transactions of the Association for Computational Linguistics},
  9:\penalty0 962--977, 2021.

\bibitem[Krause \& Guestrin(2005)Krause and Guestrin]{krause2005note}
Andreas Krause and Carlos Guestrin.
\newblock \emph{A note on the budgeted maximization of submodular functions}.
\newblock Citeseer, 2005.

\bibitem[Lewis \& Gale(1994)Lewis and Gale]{Lewis1994ASA}
David~D. Lewis and William~A. Gale.
\newblock A sequential algorithm for training text classifiers.
\newblock \emph{SIGIR}, 1994.

\bibitem[Lhoest et~al.(2021)Lhoest, Villanova~del Moral, Jernite, Thakur, von
  Platen, Patil, Chaumond, Drame, Plu, Tunstall, Davison, {\v{S}}a{\v{s}}ko,
  Chhablani, Malik, Brandeis, Le~Scao, Sanh, Xu, Patry, McMillan-Major, Schmid,
  Gugger, Delangue, Matussi{\`e}re, Debut, Bekman, Cistac, Goehringer, Mustar,
  Lagunas, Rush, and Wolf]{lhoest-etal-2021-datasets}
Quentin Lhoest, Albert Villanova~del Moral, Yacine Jernite, Abhishek Thakur,
  Patrick von Platen, Suraj Patil, Julien Chaumond, Mariama Drame, Julien Plu,
  Lewis Tunstall, Joe Davison, Mario {\v{S}}a{\v{s}}ko, Gunjan Chhablani,
  Bhavitvya Malik, Simon Brandeis, Teven Le~Scao, Victor Sanh, Canwen Xu,
  Nicolas Patry, Angelina McMillan-Major, Philipp Schmid, Sylvain Gugger,
  Cl{\'e}ment Delangue, Th{\'e}o Matussi{\`e}re, Lysandre Debut, Stas Bekman,
  Pierric Cistac, Thibault Goehringer, Victor Mustar, Fran{\c{c}}ois Lagunas,
  Alexander Rush, and Thomas Wolf.
\newblock Datasets: A community library for natural language processing.
\newblock In \emph{Proceedings of the 2021 Conference on Empirical Methods in
  Natural Language Processing: System Demonstrations}, 2021.

\bibitem[Li \& Qiu(2023)Li and Qiu]{li2023finding}
Xiaonan Li and Xipeng Qiu.
\newblock Finding supporting examples for in-context learning.
\newblock \emph{arXiv preprint arXiv:2302.13539}, 2023.

\bibitem[Liu et~al.(2021)Liu, Shen, Zhang, Dolan, Carin, and
  Chen]{liu2021makes}
Jiachang Liu, Dinghan Shen, Yizhe Zhang, Bill Dolan, Lawrence Carin, and Weizhu
  Chen.
\newblock What makes good in-context examples for gpt-$3 $?
\newblock \emph{arXiv preprint arXiv:2101.06804}, 2021.

\bibitem[Liu et~al.(2019)Liu, Ott, Goyal, Du, Joshi, Chen, Levy, Lewis,
  Zettlemoyer, and Stoyanov]{liu2019roberta}
Yinhan Liu, Myle Ott, Naman Goyal, Jingfei Du, Mandar Joshi, Danqi Chen, Omer
  Levy, Mike Lewis, Luke Zettlemoyer, and Veselin Stoyanov.
\newblock Roberta: A robustly optimized bert pretraining approach.
\newblock \emph{arXiv preprint arXiv:1907.11692}, 2019.

\bibitem[Lu et~al.(2022)Lu, Bartolo, Moore, Riedel, and
  Stenetorp]{lu2021fantastically}
Yao Lu, Max Bartolo, Alastair Moore, Sebastian Riedel, and Pontus Stenetorp.
\newblock Fantastically ordered prompts and where to find them: Overcoming
  few-shot prompt order sensitivity.
\newblock \emph{ACL}, 2022.

\bibitem[Ma et~al.(2023)Ma, Zhang, Bian, Liu, Zhang, Zhao, Zhang, Fu, Hu, and
  Wu]{ma2023fairness}
Huan Ma, Changqing Zhang, Yatao Bian, Lemao Liu, Zhirui Zhang, Peilin Zhao, Shu
  Zhang, Huazhu Fu, Qinghua Hu, and Bingzhe Wu.
\newblock Fairness-guided few-shot prompting for large language models.
\newblock \emph{arXiv preprint arXiv:2303.13217}, 2023.

\bibitem[MacQueen et~al.(1967)]{macqueen1967kmeans}
James MacQueen et~al.
\newblock Some methods for classification and analysis of multivariate
  observations.
\newblock In \emph{Proceedings of the fifth Berkeley symposium on mathematical
  statistics and probability}, volume~1, pp.\  281--297. Oakland, CA, USA,
  1967.

\bibitem[Margatina et~al.(2021)Margatina, Vernikos, Barrault, and
  Aletras]{margatina-etal-2021-active}
Katerina Margatina, Giorgos Vernikos, Lo{\"\i}c Barrault, and Nikolaos Aletras.
\newblock Active learning by acquiring contrastive examples.
\newblock In \emph{EMNLP}, 2021.

\bibitem[Margatina et~al.(2023)Margatina, Schick, Aletras, and
  Dwivedi-Yu]{margatina2023active}
Katerina Margatina, Timo Schick, Nikolaos Aletras, and Jane Dwivedi-Yu.
\newblock Active learning principles for in-context learning with large
  language models, 2023.

\bibitem[McAuley \& Leskovec(2013)McAuley and Leskovec]{mcauley2013amazon}
Julian McAuley and Jure Leskovec.
\newblock Hidden factors and hidden topics: Understanding rating dimensions
  with review text.
\newblock In \emph{Proceedings of the 7th ACM Conference on Recommender Systems
  (RecSys)}, 2013.
\newblock ISBN 9781450324090.

\bibitem[Min et~al.(2022{\natexlab{a}})Min, Lewis, Zettlemoyer, and
  Hajishirzi]{min2021metaicl}
Sewon Min, Mike Lewis, Luke Zettlemoyer, and Hannaneh Hajishirzi.
\newblock Metaicl: Learning to learn in context.
\newblock \emph{NAACL}, 2022{\natexlab{a}}.

\bibitem[Min et~al.(2022{\natexlab{b}})Min, Lyu, Holtzman, Artetxe, Lewis,
  Hajishirzi, and Zettlemoyer]{min2022rethinking}
Sewon Min, Xinxi Lyu, Ari Holtzman, Mikel Artetxe, Mike Lewis, Hannaneh
  Hajishirzi, and Luke Zettlemoyer.
\newblock Rethinking the role of demonstrations: What makes in-context learning
  work?
\newblock \emph{EMNLP}, 2022{\natexlab{b}}.

\bibitem[MosaicML(2023)]{MosaicML2023Introducing}
MosaicML.
\newblock Introducing mpt-7b: A new standard for open-source, commercially
  usable llms, 2023.
\newblock URL \url{www.mosaicml.com/blog/mpt-7b}.

\bibitem[Narayan et~al.()Narayan, Cohen, and Lapata]{narayan2018xsum}
Shashi Narayan, Shay~B. Cohen, and Mirella Lapata.
\newblock Don{'}t give me the details, just the summary! topic-aware
  convolutional neural networks for extreme summarization.
\newblock In \emph{Proceedings of the 2018 Conference on Empirical Methods in
  Natural Language Processing}.
\newblock URL \url{https://aclanthology.org/D18-1206}.

\bibitem[Nguyen \& Wong(2023)Nguyen and Wong]{nguyen2023context}
Tai Nguyen and Eric Wong.
\newblock In-context example selection with influences.
\newblock \emph{arXiv preprint arXiv:2302.11042}, 2023.

\bibitem[Olsson et~al.(2022)Olsson, Elhage, Nanda, Joseph, DasSarma, Henighan,
  Mann, Askell, Bai, Chen, et~al.]{olsson2022context}
Catherine Olsson, Nelson Elhage, Neel Nanda, Nicholas Joseph, Nova DasSarma,
  Tom Henighan, Ben Mann, Amanda Askell, Yuntao Bai, Anna Chen, et~al.
\newblock In-context learning and induction heads.
\newblock \emph{arXiv preprint arXiv:2209.11895}, 2022.

\bibitem[Penedo et~al.(2023)Penedo, Malartic, Hesslow, Cojocaru, Cappelli,
  Alobeidli, Pannier, Almazrouei, and Launay]{penedo2023falcon}
Guilherme Penedo, Quentin Malartic, Daniel Hesslow, Ruxandra Cojocaru,
  Alessandro Cappelli, Hamza Alobeidli, Baptiste Pannier, Ebtesam Almazrouei,
  and Julien Launay.
\newblock The {R}efined{W}eb dataset for {F}alcon {LLM}: outperforming curated
  corpora with web data, and web data only.
\newblock \emph{arXiv preprint arXiv:2306.01116}, 2023.

\bibitem[Perez et~al.(2021)Perez, Kiela, and Cho]{perez2021true}
Ethan Perez, Douwe Kiela, and Kyunghyun Cho.
\newblock True few-shot learning with language models.
\newblock \emph{NeurIPS}, 2021.

\bibitem[Pitis et~al.(2023)Pitis, Zhang, Wang, and Ba]{pitis2023boosted}
Silviu Pitis, Michael~R Zhang, Andrew Wang, and Jimmy Ba.
\newblock Boosted prompt ensembles for large language models.
\newblock \emph{arXiv preprint arXiv:2304.05970}, 2023.

\bibitem[Razeghi et~al.(2022)Razeghi, Logan~IV, Gardner, and
  Singh]{razeghi2022impact}
Yasaman Razeghi, Robert~L Logan~IV, Matt Gardner, and Sameer Singh.
\newblock Impact of pretraining term frequencies on few-shot numerical
  reasoning.
\newblock \emph{EMNLP-Findings}, 2022.

\bibitem[Reimers \& Gurevych(2019)Reimers and Gurevych]{reimers2019sentence}
Nils Reimers and Iryna Gurevych.
\newblock Sentence-bert: Sentence embeddings using siamese bert-networks.
\newblock \emph{EMNLP}, 2019.

\bibitem[Rubin et~al.(2022)Rubin, Herzig, and Berant]{Rubin2021LearningTR}
Ohad Rubin, Jonathan Herzig, and Jonathan Berant.
\newblock Learning to retrieve prompts for in-context learning.
\newblock \emph{NAACL}, 2022.

\bibitem[Schr{\"o}der et~al.(2022)Schr{\"o}der, Niekler, and
  Potthast]{schroder2021revisiting}
Christopher Schr{\"o}der, Andreas Niekler, and Martin Potthast.
\newblock Revisiting uncertainty-based query strategies for active learning
  with transformers.
\newblock \emph{ACL Findings}, 2022.

\bibitem[Schröder \& Niekler(2020)Schröder and Niekler]{schröder2020survey}
Christopher Schröder and Andreas Niekler.
\newblock A survey of active learning for text classification using deep neural
  networks, 2020.

\bibitem[Sener \& Savarese(2018)Sener and Savarese]{sener2018active}
Ozan Sener and Silvio Savarese.
\newblock Active learning for convolutional neural networks: A core-set
  approach.
\newblock \emph{ICLR}, 2018.

\bibitem[Settles(2009)]{settles2009active}
Burr Settles.
\newblock Active learning literature survey.
\newblock 2009.

\bibitem[Shin et~al.(2022)Shin, Lee, Ahn, Kim, Kim, Kim, Cho, Lee, Park, Ha,
  et~al.]{shin2022effect}
Seongjin Shin, Sang-Woo Lee, Hwijeen Ahn, Sungdong Kim, HyoungSeok Kim, Boseop
  Kim, Kyunghyun Cho, Gichang Lee, Woomyoung Park, Jung-Woo Ha, et~al.
\newblock On the effect of pretraining corpora on in-context learning by a
  large-scale language model.
\newblock \emph{NAACL}, 2022.

\bibitem[Shum et~al.(2023)Shum, Diao, and Zhang]{shum2023automatic}
KaShun Shum, Shizhe Diao, and Tong Zhang.
\newblock Automatic prompt augmentation and selection with chain-of-thought
  from labeled data.
\newblock \emph{arXiv preprint arXiv:2302.12822}, 2023.

\bibitem[Socher et~al.(2013)Socher, Perelygin, Wu, Chuang, Manning, Ng, and
  Potts]{socher2013sst2}
Richard Socher, Alex Perelygin, Jean Wu, Jason Chuang, Christopher~D. Manning,
  Andrew Ng, and Christopher Potts.
\newblock Recursive deep models for semantic compositionality over a sentiment
  treebank.
\newblock In \emph{Proceedings of the 2013 Conference on Empirical Methods in
  Natural Language Processing (EMNLP)}, pp.\  1631--1642, Seattle, Washington,
  USA, October 2013. Association for Computational Linguistics.
\newblock URL \url{https://www.aclweb.org/anthology/D13-1170}.

\bibitem[Su et~al.(2022)Su, Kasai, Wu, Shi, Wang, Xin, Zhang, Ostendorf,
  Zettlemoyer, Smith, et~al.]{su2022selective}
Hongjin Su, Jungo Kasai, Chen~Henry Wu, Weijia Shi, Tianlu Wang, Jiayi Xin, Rui
  Zhang, Mari Ostendorf, Luke Zettlemoyer, Noah~A Smith, et~al.
\newblock Selective annotation makes language models better few-shot learners.
\newblock \emph{ICLR}, 2022.

\bibitem[Touvron et~al.(2023)Touvron, Lavril, Izacard, Martinet, Lachaux,
  Lacroix, Rozi{\`e}re, Goyal, Hambro, Azhar, et~al.]{touvron2023llama}
Hugo Touvron, Thibaut Lavril, Gautier Izacard, Xavier Martinet, Marie-Anne
  Lachaux, Timoth{\'e}e Lacroix, Baptiste Rozi{\`e}re, Naman Goyal, Eric
  Hambro, Faisal Azhar, et~al.
\newblock Llama: Open and efficient foundation language models.
\newblock \emph{arXiv preprint arXiv:2302.13971}, 2023.

\bibitem[Vaswani et~al.(2017)Vaswani, Shazeer, Parmar, Uszkoreit, Jones, Gomez,
  Kaiser, and Polosukhin]{vaswani2017attention}
Ashish Vaswani, Noam Shazeer, Niki Parmar, Jakob Uszkoreit, Llion Jones,
  Aidan~N Gomez, {\L}ukasz Kaiser, and Illia Polosukhin.
\newblock Attention is all you need.
\newblock \emph{NeurIPS}, 2017.

\bibitem[Vazirani(2001)]{vazirani2001approximation}
Vijay~V Vazirani.
\newblock \emph{Approximation algorithms}, volume~1.
\newblock Springer, 2001.

\bibitem[Von~Oswald et~al.(2023)Von~Oswald, Niklasson, Randazzo, Sacramento,
  Mordvintsev, Zhmoginov, and Vladymyrov]{von2023transformers}
Johannes Von~Oswald, Eyvind Niklasson, Ettore Randazzo, Jo{\~a}o Sacramento,
  Alexander Mordvintsev, Andrey Zhmoginov, and Max Vladymyrov.
\newblock Transformers learn in-context by gradient descent.
\newblock In \emph{ICML}, 2023.

\bibitem[Wang \& Komatsuzaki(2021)Wang and Komatsuzaki]{wang2021gpt-j}
Ben Wang and Aran Komatsuzaki.
\newblock {GPT-J-6B: A 6 Billion Parameter Autoregressive Language Model}.
\newblock \url{https://github.com/kingoflolz/mesh-transformer-jax}, 2021.

\bibitem[Wang et~al.(2022)Wang, Min, Deng, Shen, Wu, Zettlemoyer, and
  Sun]{wang2022towards}
Boshi Wang, Sewon Min, Xiang Deng, Jiaming Shen, You Wu, Luke Zettlemoyer, and
  Huan Sun.
\newblock Towards understanding chain-of-thought prompting: An empirical study
  of what matters.
\newblock \emph{arXiv preprint arXiv:2212.10001}, 2022.

\bibitem[Wei et~al.(2022{\natexlab{a}})Wei, Tay, Bommasani, Raffel, Zoph,
  Borgeaud, Yogatama, Bosma, Zhou, Metzler, et~al.]{wei2022emergent}
Jason Wei, Yi~Tay, Rishi Bommasani, Colin Raffel, Barret Zoph, Sebastian
  Borgeaud, Dani Yogatama, Maarten Bosma, Denny Zhou, Donald Metzler, et~al.
\newblock Emergent abilities of large language models.
\newblock \emph{TMLR}, 2022{\natexlab{a}}.

\bibitem[Wei et~al.(2022{\natexlab{b}})Wei, Wang, Schuurmans, Bosma, Xia, Chi,
  Le, Zhou, et~al.]{wei2022chain}
Jason Wei, Xuezhi Wang, Dale Schuurmans, Maarten Bosma, Fei Xia, Ed~Chi, Quoc~V
  Le, Denny Zhou, et~al.
\newblock Chain-of-thought prompting elicits reasoning in large language
  models.
\newblock \emph{Advances in Neural Information Processing Systems (NeurIPS)},
  2022{\natexlab{b}}.

\bibitem[Wei et~al.(2023)Wei, Wei, Tay, Tran, Webson, Lu, Chen, Liu, Huang,
  Zhou, et~al.]{wei2023larger}
Jerry Wei, Jason Wei, Yi~Tay, Dustin Tran, Albert Webson, Yifeng Lu, Xinyun
  Chen, Hanxiao Liu, Da~Huang, Denny Zhou, et~al.
\newblock Larger language models do in-context learning differently.
\newblock \emph{arXiv preprint arXiv:2303.03846}, 2023.

\bibitem[Williams et~al.(2018)Williams, Nangia, and Bowman]{williams2018mnli}
Adina Williams, Nikita Nangia, and Samuel Bowman.
\newblock A broad-coverage challenge corpus for sentence understanding through
  inference.
\newblock In \emph{Proceedings of the 2018 Conference of the North {A}merican
  Chapter of the Association for Computational Linguistics: Human Language
  Technologies, Volume 1 (Long Papers)}, 2018.
\newblock URL \url{https://aclanthology.org/N18-1101}.

\bibitem[Wolf et~al.(2020)Wolf, Debut, Sanh, Chaumond, Delangue, Moi, Cistac,
  Rault, Louf, Funtowicz, Davison, Shleifer, von Platen, Ma, Jernite, Plu, Xu,
  Scao, Gugger, Drame, Lhoest, and Rush]{wolf-etal-2020-transformers}
Thomas Wolf, Lysandre Debut, Victor Sanh, Julien Chaumond, Clement Delangue,
  Anthony Moi, Pierric Cistac, Tim Rault, Rémi Louf, Morgan Funtowicz, Joe
  Davison, Sam Shleifer, Patrick von Platen, Clara Ma, Yacine Jernite, Julien
  Plu, Canwen Xu, Teven~Le Scao, Sylvain Gugger, Mariama Drame, Quentin Lhoest,
  and Alexander~M. Rush.
\newblock Transformers: State-of-the-art natural language processing.
\newblock In \emph{ACL (Demos)}, 2020.

\bibitem[Xie et~al.(2021)Xie, Raghunathan, Liang, and Ma]{xie2021explanation}
Sang~Michael Xie, Aditi Raghunathan, Percy Liang, and Tengyu Ma.
\newblock An explanation of in-context learning as implicit bayesian inference.
\newblock \emph{arXiv preprint arXiv:2111.02080}, 2021.

\bibitem[Xu et~al.(2023)Xu, Wang, Mao, Lyu, She, and Zhang]{xu2023knn}
Benfeng Xu, Quan Wang, Zhendong Mao, Yajuan Lyu, Qiaoqiao She, and Yongdong
  Zhang.
\newblock \$k\${NN} prompting: Beyond-context learning with calibration-free
  nearest neighbor inference.
\newblock In \emph{ICLR}, 2023.

\bibitem[Yoo et~al.(2022)Yoo, Kim, Kim, Cho, Jo, Lee, Lee, and
  Kim]{yoo2022ground}
Kang~Min Yoo, Junyeob Kim, Hyuhng~Joon Kim, Hyunsoo Cho, Hwiyeol Jo, Sang-Woo
  Lee, Sang-goo Lee, and Taeuk Kim.
\newblock Ground-truth labels matter: A deeper look into input-label
  demonstrations.
\newblock \emph{EMNLP}, 2022.

\bibitem[Zhang et~al.(2023{\natexlab{a}})Zhang, Frei, and
  Bartlett]{zhang2023trained}
Ruiqi Zhang, Spencer Frei, and Peter~L Bartlett.
\newblock Trained transformers learn linear models in-context.
\newblock \emph{arXiv preprint arXiv:2306.09927}, 2023{\natexlab{a}}.

\bibitem[Zhang et~al.(2015)Zhang, Zhao, and LeCun]{zhang2015agnews}
Xiang Zhang, Junbo Zhao, and Yann LeCun.
\newblock Character-level convolutional networks for text classification.
\newblock \emph{Advances in neural information processing systems (NeurIPS)},
  28, 2015.

\bibitem[Zhang et~al.(2022{\natexlab{a}})Zhang, Feng, and
  Tan]{zhang2022activeICL}
Yiming Zhang, Shi Feng, and Chenhao Tan.
\newblock Active example selection for in-context learning.
\newblock In \emph{EMNLP}, 2022{\natexlab{a}}.

\bibitem[Zhang et~al.(2022{\natexlab{b}})Zhang, Strubell, and
  Hovy]{zhang-etal-2022-survey}
Zhisong Zhang, Emma Strubell, and Eduard Hovy.
\newblock A survey of active learning for natural language processing.
\newblock In \emph{EMNLP}, 2022{\natexlab{b}}.

\bibitem[Zhang et~al.(2023{\natexlab{b}})Zhang, Zhang, Li, and
  Smola]{zhang2022automatic}
Zhuosheng Zhang, Aston Zhang, Mu~Li, and Alex Smola.
\newblock Automatic chain of thought prompting in large language models.
\newblock 2023{\natexlab{b}}.

\bibitem[Zhao et~al.(2021)Zhao, Wallace, Feng, Klein, and
  Singh]{zhao2021calibrate}
Zihao Zhao, Eric Wallace, Shi Feng, Dan Klein, and Sameer Singh.
\newblock Calibrate before use: Improving few-shot performance of language
  models.
\newblock In \emph{ICML}, 2021.

\end{thebibliography}
\bibliographystyle{iclr2023_conference}

\clearpage

\appendix

\section*{Appendix}

\section{\adaicl Details} \label{app:Adaicl-A}

\subsection{\adaiclb} \label{app:adaiclbase}
\Algref{alg:adaiclbase} presents the overall \adaiclb algorithm. First, \adaiclb uses the LLM's feedback to identify hard examples for the model based on their uncertainty scores. Then, it performs $k$means clustering where different clusters represent regions of hard examples and selects representative examples for each region by sampling the example closest to each of the cluster centroids. Here, the number of clusters for $k$means is $B$, so that we sample as many examples as the budget $B$ allows.

\begin{wrapfigure}{r}{0.5\textwidth} %
    \centering
    \vspace{-0.2in}
    \includegraphics[width=0.5\textwidth]{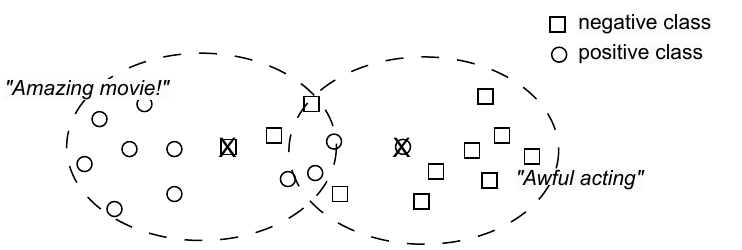}
    \vspace{-0.25in}
    \caption{Failing case for $k$means clustering.}
    \vspace{-0.17in}
    \label{fig:fail_case_clust}
\end{wrapfigure}

 \adaiclb suffers from the known limitations of $k$means clustering. It is sensitive to outlier examples, which may be selected to be annotated, assumes that the $B$ formed regions are equally important, and does not account for the effect of overlapping clusters. We provide such a failing case in \Figref{fig:fail_case_clust}. Here, the annotated examples (with $B=2$) do not effectively represent the semantic space and may not help the model understand the task.

\begin{algorithm}[H]
\centering
\caption{\adaiclb Algorithm.} \label{alg:adaiclbase}
\begin{algorithmic}[1]
   \STATE {\bfseries Input:} Model $M$, Unlabeled Set $\gU$, Budget $B$, Similarity Space $\gS$ for $k$-NN Retriever.
   \STATE {\bfseries Optional:} Initial set $\gL_0$, else $\gL_0 = \emptyset $.
   \STATE {\bfseries Hyperparameters:} threshold $\theta$.
   \STATE {\bfseries Output:} Annotated Set $\gL$.
    \newline
   
   \FOR{ $x_i \in \gU$}
    \STATE Retrieve (at most) $k$ examples from $\gL$ based on similarity $\gS$.
    \STATE Use model $M$ to obtain an uncertainty score $u_i$ for $x_i$ with $k$-shot ICL.
    \ENDFOR
    
    \STATE Determine hard set $\gU_h$ given scores $\{u_i\}$ and threshold $\theta$.
    \STATE Perform $k$means clustering of $\gU_h$ into $B$ clusters with centroids $\{ \mu_j\}_{j=1}^B$.
    \FOR{ $j=1, \dots, B$}
    \STATE $ x^\ast_i= \argmin_{x_i \in  \gU_h} || \mu_j - x_i ||$.
    \STATE Add the selected $x^\ast_i$ to $\gL=\gL \cup \{ x^\ast_i \}$ and remove it from $\gU_h=\gU_h \setminus \{ x^\ast_i \}$.
   \ENDFOR
\end{algorithmic}
\end{algorithm}

\subsection{\adaicl Algorithms} \label{app:algs}
\Algref{alg:adaicl} summarizes the overall \adaicl algorithm, and \Algref{alg:adaiclp} summarizes the overall \adaiclp algorithm. Their key differences are the following. First, \adaicl solves a \mcp problem (Line~\ref{alg:line:mcp}), while \adaicl solves a weighted \mcp problem (Line~\ref{alg:line:wmcp}). Second, \adaiclp performs a predefined number of iterations $T$ (Line~\ref{alg:line:tloop}), sampling a fixed number of examples $B_{cur} = \frac{B}{T}$ (Line~\ref{alg:line:wmcp}) per iteration. \adaicl is repeated until termination (Line~\ref{alg:line:loop}), where the number of selected examples per iteration is determined by the \mcp (Line~\ref{alg:line:mcp}) solution. 

\begin{algorithm}[H]
\centering
\caption{\adaicl Algorithm.} \label{alg:adaicl}
\begin{algorithmic}[1]
   \STATE {\bfseries Input:} Model $M$, Unlabeled Set $\gU$, Budget $B$, Similarity Space $\gS$ for $k$-NN Retriever.
   \STATE {\bfseries Optional:} Initial set $\gL_0$, else $\gL_0 = \emptyset $.
   \STATE {\bfseries Hyperparameters:} threshold $\theta$, number of neighbors $m$.
   \STATE {\bfseries Output:} Annotated Set $\gL$.
    \newline

   \STATE $B_{cur} = 0, \gL = \gL_0$.
    \STATE Create global graph $\gG_m$.
   \WHILE{$B_{cur} < B$} \label{alg:line:loop}
   \FOR{ $x_i \in \gU$}
    \STATE Retrieve (at most) $k$ examples from $\gL$ based on similarity $\gS$.
    \STATE Use model $M$ to obtain an uncertainty score $u_i$ for $x_i$ with $k$-shot ICL.
    \ENDFOR
    
    \STATE Determine hard set $\gU_h$ given scores $\{u_i\}_{i=1}^N$ and threshold $\theta$.
    \STATE Create sets $S_i$ given $\gU_h$ and $\gG_m$.
    \STATE $\{x^\ast_i\}_{i=1}^{B^\ast}$ = Greedy-\mcp\Big($\gU_h, \{S_i\}, B- B_{cur}$\Big).
    \label{alg:line:mcp}
    \STATE Add the selected $\{x^\ast_i\}_{i=1}^{B^\ast}$ to $\gL=\gL \cup \{x^\ast_i\}_{i=1}^{B^\ast}$ and remove them from $\gU=\gU \setminus \{x^\ast_i\}_{i=1}^{B^\ast}$.
    \STATE $B_{cur} = B_{cur} + B^\ast$.
   \ENDWHILE
\end{algorithmic}
\end{algorithm}

\begin{algorithm}[H]
\centering
\caption{\adaiclp Algorithm.} \label{alg:adaiclp}
\begin{algorithmic}[1]
   \STATE {\bfseries Input:} Model $M$, Unlabeled Set $\gU$, Budget $B$, Similarity Space $\gS$ for $k$-NN Retriever.
   \STATE {\bfseries Optional:} Initial set $\gL_0$, else $\gL_0 = \emptyset$.
   \STATE {\bfseries Hyperparameters:} threshold $\theta$, number of neighbors $m$, iterations $T$.
   \STATE {\bfseries Output:} Annotated Set $\gL$.
    \newline

   \STATE $B_{cur} = \frac{B}{T}, \gL = \gL_0$.
   \STATE Create global graph $\gG_m$.
   \FOR{$t \in [1, T]$} \label{alg:line:tloop}
   \FOR{ $x_i \in \gU$}
    \STATE Retrieve (at most) $k$ examples from $\gL$ based on similarity $\gS$.
    \STATE Use model $M$ to obtain an uncertainty score $u_i$ for $x_i$ with $k$-shot ICL.
    \ENDFOR
    
    \STATE Determine hard set $\gU_h$ given scores $\{u_i\}_{i=1}^N$ and threshold $\theta$.
    \STATE Create sets $S_i$ given $\gU_h$ and $\gG_m$.
    \STATE $\{x^\ast_i\}_{i=1}^{B_{cur}}$ = Greedy-weighted-\mcp\Big($\gU_h, \{S_i\}, B_{cur}$\Big). \label{alg:line:wmcp}
    \STATE Add the selected $\{x^\ast_i\}_{i=1}^{B_{cur}}$ to $\gL=\gL \cup \{x^\ast_i\}_{i=1}^{B_{cur}}$ and remove them from $\gU=\gU \setminus \{x^\ast_i\}_{i=1}^{B_{cur}}$.
   \ENDFOR
\end{algorithmic}
\end{algorithm}

\begin{wrapfigure}{r}{0.5\textwidth} %
    \centering
    \vspace{-0.2in}
    \includegraphics[width=0.5\textwidth]{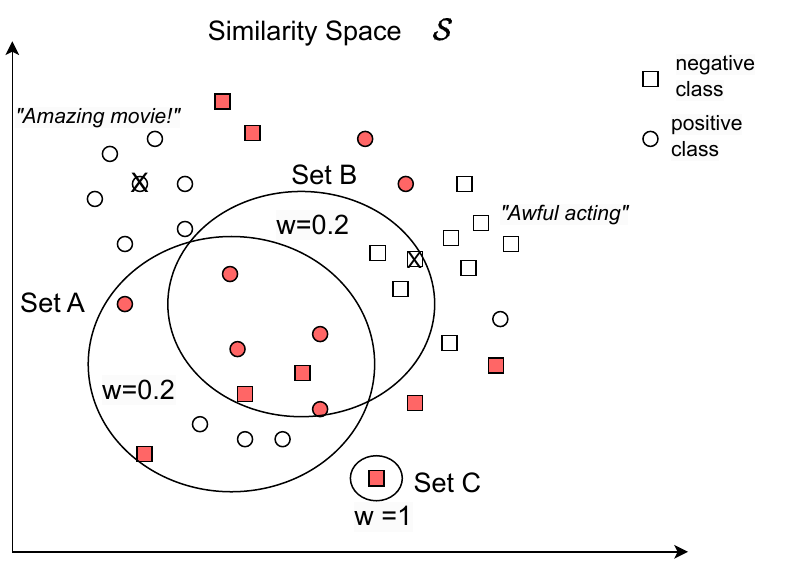}
    \vspace{-0.25in}
    \caption{A beneficial case for \adaiclp.}
    \vspace{-0.17in}
    \label{fig:adaiclp_case}
\end{wrapfigure}
We provide an example of \adaiclp's advantage in \Figref{fig:adaiclp_case}. In this example, \adaiclp's re-weighting schema scores Sets A and B higher than Set C, which contains only a single hard example. On the contrary, if Set A is selected by \adaicl, all the examples of Set B would be marked as covered resulting in a zero total score. The next best scoring set would be Set C, which does not effectively represent the hardness of the examples. In \adaiclp, Set B is scored higher than Set C due to its new weight $w=0.2$.

\subsection{Set Construction} \label{app:set}
In this work, we represent the region $S_i$ around each example $x_i$ as its egonet. Initially, we build a global graph $\gG_m$ as the $m$-nearest neighbors graph. The nearest neighbors are obtained with similarity metrics based on the space $\gS$, i.e., via cosine similarity.  For example, we can have the following edge sets for two nodes $v_1$ and $v_2$ with $m=4$:
$
     \{ v_1 \xrightarrow[]{} v_2, \; v_1 \xrightarrow[]{} v_3, \; v_1 \xrightarrow[]{} v_4, \; v_1 \xrightarrow[]{} v_5 \}, \text{ and } 
     \{ v_2 \xrightarrow[]{} v_3, \; v_2 \xrightarrow[]{} v_6, \; v_2 \xrightarrow[]{} v_7, \; v_2 \xrightarrow[]{} v_1 \}.
$

Deriving sets $S_i$ from the global graph $\gG_m$ depends on the LLM's feedback as we are interested in hard examples $x_i \in \gU_h$ for the model. Thus, we \emph{color} each node as a hard or an easy node based on whether they belong to $\gU_h$. For example, we can have $\gU_h = \{ v_1, v_2, v_3, v_4\}$ and $\{v_5, v_6, v_7\} \notin U_h$ for the case above.

For every hard node $v \in \gU_h$, we construct its 1-hop egonet. We consider edges that \emph{direct towards} $v$ from other hard nodes $v' \in \gU_h$. This ensures that representative examples $x_i$, that are likely to be retrieved during ICL inference, have denser egonets. For example, if we have $v_2 \xrightarrow[]{} v_1, v_3 \xrightarrow[]{} v_1$ (we exclude links from easy nodes, e.g., $v_5 \xrightarrow[]{} v_1$), we obtain $S_1 = \text{egonet}(v_1) = \{v_2, v_3\}$. Similarly, we obtain egonets for other nodes, e.g., $S_2 = \text{egonet}(v_2) = \{v_1, v_4\}$, while some nodes might have empty egonets if only easy nodes directs towards them, e.g., $S_3 = \text{egonet}(v_3) =\emptyset$. We experiment with both 1-hop and 2-hop set constructions. In the latter case, each node $v$ is represented by its egonet along with union of the egonets of its neighbors, e.g., $S^{(2)}_1 = \text{ego}(v_1) \cup \{ \text{ego}(v'): v'  \in \text{ego}(v_1) \} = \{ v_2, v_3\}\cup \text{ego}(v_2) \cup  \text{ego}(v_3) = \{v_2, v_3, v_4\}.$

One hyperparameter that controls the quality of the generated egonets is $m$, which is used during the construction of graph $\gG_m$. In order to determine $m$, we employ a heuristic rule based on the desired maximum iterations $\hat{T}$ until the budget $B$ is exhausted, as well as the minimum number of hard examples $N_{\hat{\theta}}$ to be covered at each iteration, where $N_{\hat{\theta}} = \lfloor \hat{\theta} N_{\theta} \rfloor$ and $\hat{\theta} \in [0,1]$ is a hyper-parameter with default value $\hat{\theta} = 0.5$. Assuming the graph has reciprocal edges, each node has approximately $\theta m$  and $\theta^2 m^2$ hard examples as neighbors for 1-hop and 2-hop sets, respectively. If at each iteration we annotate $\frac{B}{\hat{T}}$ examples, and we wish to cover at least $N_{\hat{\theta}}$ hard examples, we need to satisfy $N_{\hat{\theta}} \leq \frac{B}{\hat{T}} \theta m$ (for 1-hop sets) and $N_{\hat{\theta}} \leq \frac{B}{\hat{T}} \theta^2 m^2$ (for 2-hop sets). Thus, the heuristic-based rule is given by 
\begin{equation}
  \begin{cases}
    \frac{\hat{T} N_{\hat{\theta}}}{\theta B} \leq m \leq  \frac{\hat{T} N_{\theta}}{\theta B} & \text{for 1-hop sets},  \\
    \frac{\hat{T} N_{\hat{\theta}}}{\theta^2 B} \leq m^2 \leq  \frac{\hat{T} N_{\theta}}{\theta^2 B} & \text{for 2-hop sets}, 
  \end{cases}
  \label{eq:heuristic}
\end{equation}
which is adjustable to the portion of the examples that we account as hard ones (the right hand side is derived due to constraint of maximum iterations $\hat{T}$).  Moreover, instead of having a $m$-nn graph $\gG_m$, we experiment with a threshold-based $\delta$-graph $\gG_{\delta}$, where we set the threshold accordingly.

\section{Experimental Setting Details}  \label{app:Setting-B}

\subsection{Datasets}  \label{app:data}

We performed empirical evaluation  with nine NLP datasets that cover well-studied tasks, such as topic classification (AGNews~\citep{zhang2015agnews}, TREC~\citep{hovy2001trec}), sentiment analysis (SST2~\citep{socher2013sst2}, Amazon~\citep{mcauley2013amazon}), natural language inference (RTE~\citep{bentivogli2009rte}, MRPC~\citep{dolan2004mrpc}, MNLI~\citep{williams2018mnli}), text summarization (XSUM~\citep{narayan2018xsum}) and math reasoning (GSM8K~\citep{cobbe2021gsm8k}). We provide examples of these datasets in Table~\ref{tab:Datasets}, which we access via Hugging Face package~\citep{lhoest-etal-2021-datasets}.

\begin{table}[h]
\centering
\caption{Dataset examples. $<S_1>$ denotes the input sequences. }
\label{tab:Datasets}%
\resizebox{\linewidth}{!}{
\begin{threeparttable}
    \begin{tabular}{p{1.2cm}|p{2cm}|p{7cm}|p{4cm}}
        \toprule
         \textbf{Dataset} & \textbf{Task} & \textbf{Example} $x$ & \textbf{Labels/Annotations} $y$ \\
         \midrule
        AGNews & Topic Classification & $<S_1>$: ``Amazon Updates Web Services Tools, Adds Alexa Access The Amazon Web Services (AWS) division of online retail giant Amazon.com yesterday released Amazon E-Commerce Service 4.0 and the beta version of Alexa Web Information Service.'' & World, Sport, Business, \underline{Sci-Tech} \\
        \hline
        TREC & Answer Type Classification & $<S_1>$: ``What is the date of Boxing Day?'' & Abbreviation, Entity, Description, Human, Location, \underline{Numeric} \\
        \midrule
        \midrule
        SST2 & Sentiment Analysis & $<S_1>$: ``covers this territory with wit and originality , suggesting that with his fourth feature'' & \underline{Positive}, Negative \\
        \hline
        Amazon & Sentiment Analysis & $<S_{1a}>$:``Very Not Worth Your Time'', $<S_{1b}>$:``The book was written very horribly. I would never in my life recommend such a book..." & Positive, \underline{Negative} \\ 
        \midrule
        \midrule
        RTE & Natural Language Inference & $<S_1>$:``In a bowl, whisk together the eggs and sugar until completely blended and frothy.'', $<S_2>$:``In a bowl, whisk together the egg, sugar and vanilla until light in color.'' & Entailment, \underline{Not Entailment} \\
        \hline
        MRPC &Paraphrase Detection & $<S_1>$:``He said the foodservice pie business doesn't fit the company's long-term growth strategy.'', $<S_2>$:``The foodservice pie business does not fit our long-term growth strategy.'' & \underline{Equivalent}, Not Equivalent \\
        \hline
        MNLI & Natural Language Inference& $<S_1>$:``The new rights are nice enough'', $<S_2>$: ``Everyone really likes the newest benefits'' & Entailment, \underline{Neutral}, Contradiction \\
        \midrule 
        \midrule 
        XSUM & Summarization & $<S_1>$:``The 3kg (6.6lb) dog is set to become part of a search-and-rescue team used for disasters such as earthquakes. Its small size means it will be able to squeeze into places too narrow for dogs such as German Shepherds. Chihuahuas, named after a Mexican state, are one of the the smallest breeds of dog. "It's quite rare for us to have a chihuahua work as a police dog (said a police spokeswoman in Nara, western Japan). We would like it to work hard by taking advantage of its small size. Momo, aged seven, will begin work in January.'' & ``A chihuahua named Momo (Peach) has passed the exam to become a dog in the police force in western Japan, in what seems to be a first.'' \\
        \hline
        GSM8K & Math Reasoning & $<S_1>$:``James writes a 3-page letter to 2 different friends twice a week. How many pages does he write a year?'' & ``He writes each friend 3*2=6 pages a week So he writes 6*2=12 pages every week That means he writes 12*52=624 pages a year. Thus, the answer is 624.'' \\
        \bottomrule
    \end{tabular}%
    \end{threeparttable}
}

\end{table}%

Each dataset contains official train/dev/test splits. We follow Vote$k$ and sample 256 examples randomly from the test set (if it is publicly available, otherwise from the dev set) as test data. For the train data, we remove the annotations before our active learning setup. As it is infeasible to evaluate the LLM's feedback on all instances due to computational constraints, e.g., Amazon dataset has more than 1 million instances, we randomly subsample 3,000 instances, which we cluster into 310 groups, and we select the 310 examples closest to the centroids as candidate examples for annotation. We repeat the above processes for both the train and test sets three times with different seeds and report mean and standard deviation results. In transductive settings, we only consider the test data for annotation. In this case, we evaluate performance on all test examples, but we also exclude retrieving examples that lead to self-label leakage issues.

\subsection{Prompts} \label{app:prompt-data}

As a design choice of the input prompts, we slightly modify the templates proposed by~\citet{gao-etal-2021-making-prompt} to transform them as a continuation task. We find that these are more challenging prompts for the large LMs,  which we present in Table~\ref{tab:prompts} (top). In \Secref{app:prompt}, we also experiment with alternative prompt templates similar to~\citet{su2022selective}, as shown in  Table~\ref{tab:prompts} (bottom).

\begin{table}[h]
\centering
\caption{Prompt templates for the ICL demonstrations. }
\label{tab:prompts}%
\resizebox{\linewidth}{!}{
\begin{threeparttable}
    \begin{tabular}{lll}
        \toprule
         \textbf{Task} & \textbf{Template} & \textbf{Continuation (label word)}  \\
         \midrule
         & \textbf{Default}  & \\
        AGNews & Content: $<S_1> \backslash n$ &  World, Sport, Business, Sci-Tech \\
        TREC & Content: $<S_1> \backslash n$ &  Abbreviation, Entity, Description, Human, Location, Numeric \\
        SST2 & $ <S_1>$. It was & great, terrible\\
        Amazon & $ <S_{1a}> <S_{1b}>$. It was & great, terrible\\
        RTE & $ <S_1>$? [MASK], $ <S_2>$ & [MASK]: Yes, [MASK]: No\\
        MRPC & $ <S_1>$? [MASK], $ <S_2>$ & [MASK]: Yes, [MASK]: No\\
        MNLI & $ <S_1>$? [MASK], $ <S_2>$ & [MASK]: Yes, [MASK]: Maybe, [MASK]: No\\
        \midrule
        & \textbf{Alternative} & \\
        AGNews & Content: $<S_1>$ Topic: &  World, Sport, Business, Sci-Tech \\
        TREC & Content: $<S_1>$ Answer Type: &  Abbreviation, Entity, Description, Human, Location, Numeric \\
        SST2 &  Content: $ <S_1>$ Sentiment: & Positive, Negative\\
        Amazon & Title: $<S_{1a}>$ Review: $<S_{1b}>$ Sentiment: & Positive, Negative\\
        RTE & $ <S_1>$. Question: $ <S_2>$. True or False? Answer: & True, False\\
        MRPC & Are the following sentences equivalent or not equivalent? $ <S_1> \backslash n <S_2>$ & equivalent, not equivalent\\
        MNLI & $<S_1>$. Based on that information, is the claim $<S_2>$ True, False, or Inconclusive? Answer: & True, Inconclusive, False\\
        \bottomrule
    \end{tabular}%
    \end{threeparttable}
}

\end{table}%

\subsection{Configurations} \label{app:conf}
As summarized in \Figref{fig:setting}, the design space includes the unlabeled set $\gU$, the number of ICL examples $k$, the similarity space $\gS$, the budget $B$, and the LLM $M$. We use the default hyper-parameters of the Transformers library~\citep{wolf-etal-2020-transformers}  for each LLM. We obtain the initial pool of annotated examples $\gL_0$ via $k$means so that we reduce randomness.
We summarize the experimental configurations in Table~\ref{tab:setting-all}.

\begin{table}[h]
\centering
\caption{Experimental setting configurations. }
\label{tab:setting-all}%
\resizebox{\linewidth}{!}{
\begin{threeparttable}
    \begin{tabular}{l|cccccc}
        \toprule
         Setting & Models $M$ & Train/Test $\gU$ & Budget $B$ & $k$-shot & Retriever, $\gS$ & Init. \\
         \toprule
         \textbf{Main} & \\
        \Figref{fig:main} & GPTJ, GPT-Neo & Inductive & 20 & 5 & SBERT & $|\gL_0|=10$ \\
        
        XSUM & Falcon-40B, LLaMa-65B & Transductive & 10 & Context-limit & SBERT & Zero-shot \\
        GSM8K & Falcon-40B, LLaMa-65B & Transductive & 20 & 5 & BERT,SBERT & Zero-shot \\
        Table~\ref{tab:kshot} & GPT-J, MPT & Transductive & 5,10 & 5,10 & SBERT & Zero-shot\\
        Table~\ref{tab:sbert} & GPT-Neo & Inductive & 20 & 5 & SBERT, RoBERTa, BERT & $|\gL_0|=10$\\
        
        \Figref{fig:phases} & GPT-Neo & Transductive & 0-45 & Context-limit  & SBERT & Zero-shot \\
        
         \Figref{fig:llms} & GPT-J, MPT, Falcon, LLaMa (6-7B) & Transductive & 0-20 & Context-limit  & SBERT & Zero-shot \\
         \Figref{fig:calibr} & GPT-Neo & Transductive & 0-45 & Context-limit  & SBERT & Zero-shot \\
         \midrule
         Appendices \ref{app:Results-C}, \ref{app:Abla-D} & GPT-J, GPT-Neo & Inductive & 20 & 5 & SBERT & $|\gL_0|=10$ \\
        \bottomrule
    \end{tabular}%
    \begin{tablenotes}
    \item Context-limit means that we retrieve as many few-shot examples as the input token-length limit allows. For example, XSUM has long sequences, where we usually have 3-shot examples, while for TREC  we can use as many as 80-shot examples.
    \end{tablenotes}
    \end{threeparttable}
}

\end{table}%

\section{Further Results} \label{app:Results-C}

\subsection{Base Results (Full)} \label{app:full_res}
Tables~\ref{tab:gptj} and ~\ref{tab:gptneo} give the full results of \Figref{fig:main} for GPT-J and GPT-Neo, accordingly. \adaicl performs the best over all tasks, while the second-best method is \adaiclb. 

\begin{table}[h]
\centering
\caption{Performance comparison for GPT-J (6B).}
\label{tab:gptj}%
\resizebox{\linewidth}{!}{
\begin{threeparttable}
    \begin{tabular}{l|rr|rr|rrr}
        \toprule
         & \multicolumn{2}{c|}{\textbf{Topic Classification}} & \multicolumn{2}{c|}{\textbf{Sentiment Analysis}} & \multicolumn{3}{c}{\textbf{Natural Language Inference}} \\
        & \multicolumn{1}{c}{AGNews} & \multicolumn{1}{c|}{TREC} & \multicolumn{1}{c}{SST2} & \multicolumn{1}{c|}{Amazon} &\multicolumn{1}{c}{RTE} & \multicolumn{1}{c}{MRPC} & \multicolumn{1}{c}{MNLI} \\
        \midrule
        
        Random &  $68.87_{\pm 5.39}$ & $49.34_{\pm 3.19}$ & $81.63_{\pm 0.30}$ & $87.89_{\pm 1.77}$ &
        $52.86_{\pm 2.41}$ & $69.01_{\pm 4.61}$ & $39.58_{\pm 3.98}$\\
        Fast-vote$k$ & $73.69_{\pm 2.39}$ & $49.61_{\pm 4.43}$ & $78.99_{\pm 4.53}$ & $89.58_{\pm 0.80}$ & $53.00_{\pm 0.49}$ & $68.23_{\pm 2.89}$ & $39.97_{\pm 3.98}$ \\
        Vote$k$ & $72.26_{\pm 1.27}$ & $45.83_{\pm 1.75}$ & $80.45_{\pm 1.47}$ & $85.80_{\pm 3.80}$ & $54.16_{\pm 2.30}$ & $68.10_{\pm 2.75}$ & $39.72_{\pm 2.07}$ \\ 
        Hardest & $72.13_{\pm 2.12}$ & $35.93_{\pm 5.53}$ & $82.67_{\pm 1.64}$ & $87.36_{\pm 0.66}$ & $55.33_{\pm 2.25}$ & $66.80_{\pm 5.25}$ & $38.80_{\pm 1.11}$ \\ 
        \adaiclb & $73.56_{\pm 2.96}$ & $50.64_{\pm 9.11}$ & $84.11_{\pm 3.25}$ & $91.01_{\pm 1.77}$ & $	52.73_{\pm 2.21}$ & $66.53_{\pm 4.78}$ & $38.66_{\pm 4.50}$ \\

        Best (Avg.) & \multicolumn{2}{c}{\adaiclb ($62.10$)} & \multicolumn{2}{c}{\adaiclb ($87.56$) } & \multicolumn{3}{c}{Vote$k$ ($53.99$)} \\
        \midrule
       \adaicl & $76.89_{\pm 3.01}$ & $51.95_{\pm 8.43}$ & $82.81_{\pm 1.39}$ & $90.49_{\pm 1.57}$ & $56.90_{\pm  1.75}$ & $70.17_{\pm 1.72}$ & $40.36_{\pm 1.75}$ \\
       \adaiclp & $77.08_{\pm 1.11}$ & $53.38_{\pm  5.10}$ & $84.24 _{\pm 1.32}$ & $92.45_{\pm 1.50}$ & $55.07_{\pm  0.85}$ & $68.49_{\pm 0.97}$ & $36.58_{\pm 1.12}$ \\
       Best (Avg.) & \multicolumn{2}{c}{ \adaiclp ($65.23$)} & \multicolumn{2}{c}{\adaiclp ($88.35$)} & \multicolumn{3}{c}{\adaicl ($55.81$)} \\
       \midrule
       $\Delta$-Gain (Absolute) & \multicolumn{2}{c}{$+3.13$} & \multicolumn{2}{c}{$+0.79$} & \multicolumn{3}{c}{$+1.82$}\\
        \bottomrule
    \end{tabular}%
    \end{threeparttable}
}

\end{table}%

\begin{table}[h]
\centering
\caption{Performance comparison for GPT-Neo (1.3B).}
\label{tab:gptneo}%
\resizebox{\linewidth}{!}{
\begin{threeparttable}
    \begin{tabular}{l|rr|rr|rrr}
        \toprule
         & \multicolumn{2}{c|}{\textbf{Topic Classification}} & \multicolumn{2}{c|}{\textbf{Sentiment Analysis}} & \multicolumn{3}{c}{\textbf{Natural Language Inference}} \\
        & \multicolumn{1}{c}{AGNews} & \multicolumn{1}{c|}{TREC} & \multicolumn{1}{c}{SST2} & \multicolumn{1}{c|}{Amazon} &\multicolumn{1}{c}{RTE} & \multicolumn{1}{c}{MRPC} & \multicolumn{1}{c}{MNLI} \\
        \midrule
        
        Random &  $59.47_{\pm 8.54}$ & $54.68_{\pm 1.68}$ & $	68.48_{\pm 1.87}$ & $73.95_{\pm 2.03}$ &
        $48.30_{\pm 1.30}$ & $64.48_{\pm 7.67}$ & $40.99_{\pm 0.97}$\\
        
        Fast-vote$k$ & $62.23_{\pm 3.89}$ & $46.48_{\pm 3.04}$ & $	69.78_{\pm 8.34}$ & $69.39_{\pm 0.98}$ & $50.64_{\pm 1.02}$ & $64.19_{\pm 0.97}$ & $38.40_{\pm 0.92}$ \\
        
        Vote$k$ & $62.77_{\pm 4.82}$ & $53.12_{\pm 4.07}$ & $73.69_{\pm 9.05}$ & $75.13_{\pm 0.98}$ & $49.99_{\pm 0.32}$ & $67.44_{\pm 2.96}$ & $39.18_{\pm 1.60}$ \\        

        Hardest & $65.10_{\pm 2.43}$ & $49.34_{\pm 2.17}$ & $71.48_{\pm 5.32}$ & $75.00_{\pm 2.49}$ & $52.86_{\pm 0.80}$ & $61.84_{\pm 4.79}$ & $37.49_{\pm 1.77}$ \\ 
        
        \adaiclb & $70.17_{\pm 1.84}$ & $48.24_{\pm 0.98}$ & $77.86_{\pm 1.02}$ & $75.77_{\pm 3.62}$ & $	53.77_{\pm 0.73}$ & $64.71_{\pm 7.39}$ & $39.71_{\pm 1.03}$ \\

        Best (Avg.) & \multicolumn{2}{c}{\adaiclb ($59.21$)} & \multicolumn{2}{c}{\adaiclb ($76.82$)} & \multicolumn{3}{c}{Vote$k$ ($52.20$)} \\
        \midrule
        
       \adaicl & $70.95_{\pm 1.87}$ & $55.33_{\pm 2.57}$ & $79.68_{\pm 1.77}$ & $77.73_{\pm 2.23}$ & $53.12_{\pm  1.59}$ & $67.05_{\pm 8.10}$ & $42.96_{\pm 2.92}$ \\
       
       \adaiclp & $69.39_{\pm 1.35}$ & $59.89_{\pm  2.07}$ & $79.03 _{\pm 2.47}$ & $77.08_{\pm 1.50}$ & $51.16_{\pm 1.39}$ & $	65.69_{\pm 8.92}$ & $40.49_{\pm 2.04}$ \\
       Best (Avg.) & \multicolumn{2}{c}{ \adaiclp ($64.64$)} & \multicolumn{2}{c}{\adaicl ($78.71$)} & \multicolumn{3}{c}{\adaicl ($54.38$)} \\
       \midrule
       $\Delta$-Gain (Absolute) & \multicolumn{2}{c}{$+5.53$} & \multicolumn{2}{c}{$+1.99$} & \multicolumn{3}{c}{$+2.28$}\\
        \bottomrule
    \end{tabular}%
    \end{threeparttable}
}

\end{table}%

\subsection{Retriever Ablation} \label{app:retriever}

\begin{table}[tb]
\centering
\caption{Performance comparison across different retrieval and semantic similarity configurations.}
\label{tab:sbert_all}%
\resizebox{\linewidth}{!}{
\begin{threeparttable}
    \begin{tabular}{l|ccc|ccc|ccc|c}
        \toprule
         Retriever, $\gS \xrightarrow{}$ & \multicolumn{3}{c|}{SBERT-all-mpnet-base} & \multicolumn{3}{c|}{RoBERTa-nli-large-mean-tokens} & \multicolumn{3}{c|}{BERT-nli-large-cls-pool} & Avg. \\
        & TREC & SST2 & Amazon & TREC & SST2 & Amazon & TREC & SST2 & Amazon &  \\
        \midrule
        Random &  $54.68_{\pm 1.68}$ & $68.48_{\pm 1.87}$ & $73.95_{\pm 2.03}$ & $37.23_{\pm 2.30}$ & $74.21_{\pm 3.50}$ & $84.46_{\pm 3.21}$ & $34.75_{\pm 2.41}$ & $72.65_{\pm 5.82}$ & $80.20_{\pm 3.34}$ & $64.51$\\
        Vote$k$ & $54.81_{\pm 0.49}$ & $73.69_{\pm 9.05}$ & $ 75.13_{\pm 0.98}$ & $37.77_{\pm 4.65}$ &$76.16_{\pm 2.23}$ & $84.11_{\pm 1.28}$& $42.43_{\pm 3.34}$ & $80.85_{\pm 2.09}$& $83.59_{\pm 1.77}$ & $67.61$\\
        \adaiclb &  $48.24_{\pm 0.98}$ & $77.86_{\pm 1.02}$ & $75.77_{\pm 3.63}$ & $38.12_{\pm 5.74}$ & $78.12_{\pm 5.30}$ & $\underline{85.93}_{\pm 2.30}$ & $	38.15_{\pm 3.10}$ & $78.64_{\pm 2.78}$ & $\underline{85.80}_{\pm 1.75}$ & $67.40$\\ 
       \textbf{\adaicl} &    $\underline{55.33}_{\pm 2.57}$ & $\underline{79.68}_{\pm 2.47}$ & $\underline{77.73}_{\pm 2.23}$ & $\underline{39.06}_{\pm 3.37}$ & $\underline{81.11}_{\pm 1.50}$ & $ 85.15_{\pm 0.55}$ & $44.06_{\pm 2.49}$ & $\underline{80.85}_{\pm 2.83}$ & $84.65_{\pm 3.52}$ & $\textbf{69.74}$\\
        \midrule
        \textit{\underline{random reorder}} & \\
        Random &  $47.39_{\pm 2.89}$ & $66.57_{\pm 3.86}$ & $76.42_{\pm 2.57}$ & $31.87_{\pm 4.45}$ & $69.74_{\pm 5.84}$ & $	81.24_{\pm 2.72}$ & $40.34_{\pm 1.58}$ & $76.55_{\pm 4.06}$ & $79.54_{\pm 2.64}$ & $63.29$\\
        Vote$k$ & $45.43_{\pm 2.94}$ & $72.00_{\pm 3.73}$ & $73.04_{\pm 1.77}$ & $37.62_{\pm 5.15}$ &$75.91_{\pm 5.58}$ & $	82.03_{\pm 0.84}$& $	39.57_{\pm 3.19}$ & $79.29_{\pm 2.55}$& $81.24_{\pm 1.43}$ & $65.13$\\
        \adaiclb &  $50.64_{\pm 5.43}$ & $74.99_{\pm 6.37}$ & $74.86_{\pm 4.65}$ & $36.56_{\pm 2.98}$ & $75.77_{\pm 0.63}$ & $	84.89_{\pm 2.02}$ & $\underline{45.56}_{\pm 3.01}$ & $79.55_{\pm 0.18}$ & $85.41_{\pm 0.91}$ & $67.58$ \\ 
       \textbf{\adaicl} & $52.22_{\pm 5.19}$ & $78.12_{\pm 4.71}$ & $75.64_{\pm 3.51}$ & $38.93_{\pm 0.48}$ & $76.68_{\pm 2.71}$ & $ 84.11_{\pm 1.43}$ & $44.00_{\pm 1.63}$ & $78.51_{\pm 4.52}$ & $85.28_{\pm 4.65}$ & $\textbf{68.16}$\\
        \bottomrule
    \end{tabular}%
    \end{threeparttable}
}

\end{table}%

A benefit of the $k$-NN retriever is that it can determine the order of the input few-shot examples by semantic similarity scores. In general, we place demonstrations with higher similarity closer to the test instance. Table~\ref{tab:sbert_all} gathers results for different retrievers as well as when we randomly re-order the input demonstrations. \adaicl is the most robust method and outperforms other baselines regardless of the choice of the retriever.

\subsection{Prompt Template Ablation} \label{app:prompt}

\begin{table}[h]
\centering
\caption{Prompt template ablation study.}
\label{tab:prompt-abla}%
\resizebox{\linewidth}{!}{
\begin{threeparttable}
    \begin{tabular}{l|rrrr|rrr}
        \toprule
         & \multicolumn{4}{c|}{GPT-Neo} & \multicolumn{3}{c}{GPT-J} \\
        & \multicolumn{1}{c}{AGNews} & \multicolumn{1}{c}{TREC} & \multicolumn{1}{c}{SST2} & \multicolumn{1}{c|}{Amazon} &\multicolumn{1}{c}{RTE} & \multicolumn{1}{c}{MRPC} & \multicolumn{1}{c}{MNLI} \\
        \midrule
         & \multicolumn{7}{c}{\textbf{Default Prompts}}  \\
        Random &  $59.47_{\pm 8.54}$ & $54.68_{\pm 1.68}$ & $	68.48_{\pm 1.87}$ & $73.95_{\pm 2.03}$ &
        $52.86_{\pm 2.41}$ & $69.01_{\pm 4.61}$ & $39.58_{\pm 3.98}$\\
        Vote$k$ & $62.77_{\pm 4.82}$ & $53.12_{\pm 4.07}$ & $73.69_{\pm 9.05}$ & $75.13_{\pm 0.98}$ & $54.16_{\pm 2.30}$ & $68.10_{\pm 2.75}$ & $39.72_{\pm 2.07}$ \\ 
        
       \adaicl & $70.95_{\pm 1.87}$ & $55.33_{\pm 2.57}$ & $79.68_{\pm 1.77}$ & $77.73_{\pm 2.23}$ & $56.90_{\pm  1.75}$ & $70.17_{\pm 1.72}$ & $40.36_{\pm 1.75}$ \\
       
       \midrule
        & \multicolumn{7}{c}{\textbf{Alternative Prompts}} \\

        Random &  $73.69_{\pm 1.21}$ & $51.76_{\pm 4.55}$ & $59.89_{\pm 3.98} $ & $73.82_{\pm 3.35}$ &
        $56.41_{\pm2.13}$ & $ 56.37_{\pm 3.72}$ & $ 38.93_{\pm 1.18}$\\
        Vote$k$ & $	72.78_{\pm 2.12}$ & $50.38_{\pm 5.90}$ & $64.84_{\pm 2.92}$ & $73.43_{\pm 2.23}$ & 
        $56.38_{\pm 2.70}$ & $51.95_{\pm 2.53}$ & $ 40.49_{\pm 2.05}$ \\ 
        
       \adaicl & $76.95_{\pm 1.27}$ & $54.94_{\pm 1.43}$ & $65.88_{\pm 4.58} $ & $75.64_{\pm 1.29}$ & 
       $56.37_{\pm 1.29}$ & $59.22_{\pm 2.39}$ & $35.40_{\pm 1.31}$ \\
       
        \bottomrule
    \end{tabular}%

    \end{threeparttable}
}

\end{table}%

Table~\ref{tab:prompt-abla} reports results when we use alternative ICL prompt templates (Table~\ref{tab:prompts}) for the input examples. \adaicl is robust to the design of the prompt templates, where it outperforms other baselines in most datasets.

\subsection{Calibration Analysis via Simplices} \label{app:calibr2}

Calibration has recently also been studied from the perspective of simplices \citep{heese2023calibrated}. Abstractly, a simplex is the generalization of the notion of a triangle/ tetrahedron to arbitrary dimensions. Here, we present a simplified study of calibration via the lens of simplices, where we test if the predicted label of the test instance lies within the simplex given by the retrieved examples \emph{having the same label}. 
In more detail, given a prediction $y_{test}=\bm{y}$ for test instance $x_{test}$, we first obtain the subset of retrieved examples for $x_{test}$, i.e. $\gX_{subset} \subseteq \gX_{retrieved} \subset B$ which share the same label $\bm{y}$. We then construct a simplex using the SBERT embeddings of $\gX_{subset}$, run PCA to obtain low dimensional embeddings and test to see if the low dimensional embedding of $x_{test}$ lies within the above constructed simplex. 

\begin{figure*}[t]
    \centering
    \includegraphics[width=1.0\linewidth]{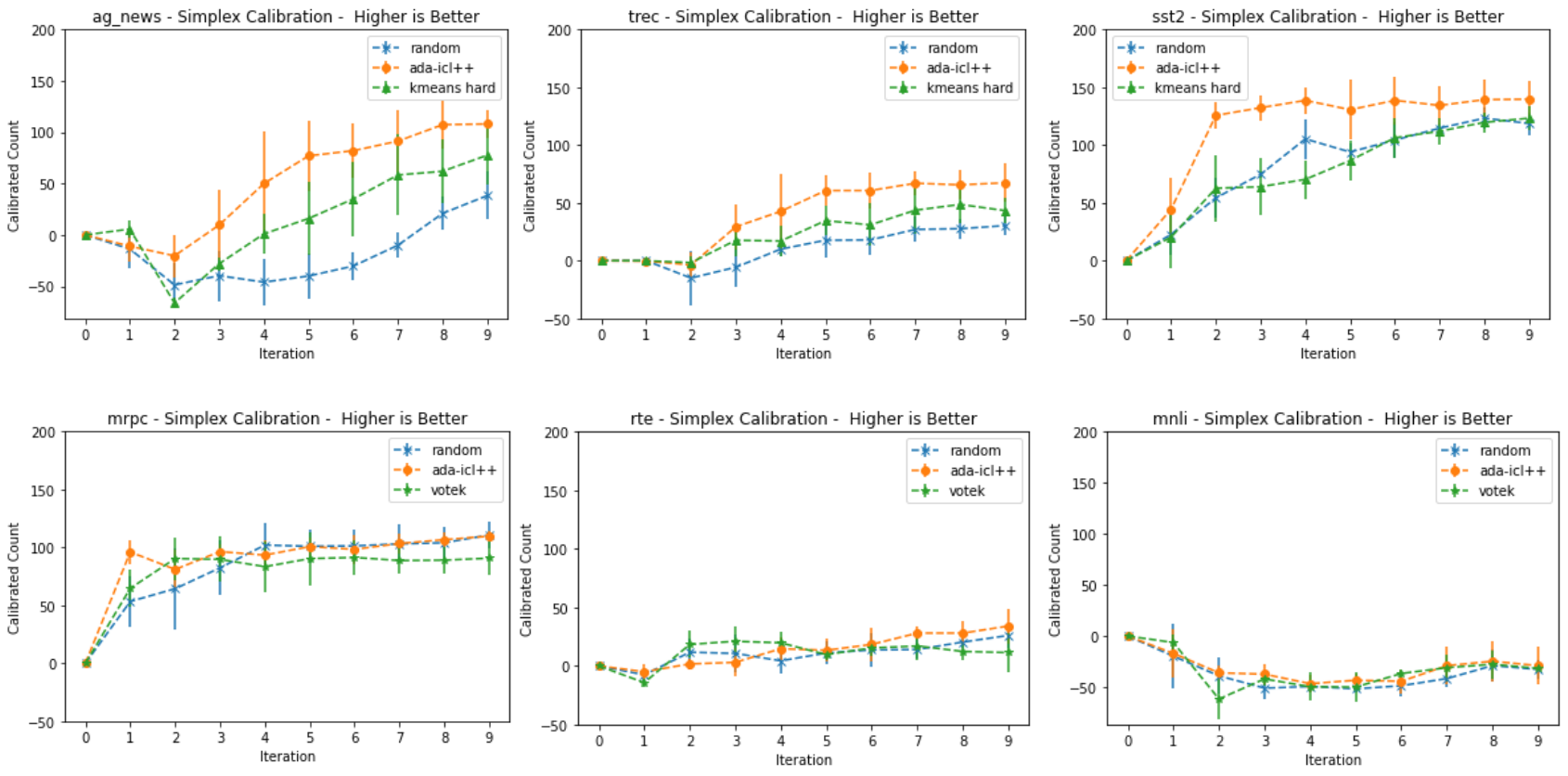}
    \caption{Calibration analysis via simplicies where we compare \adaicl against the random selection algorithm and the best performing baseline for each dataset.}
    \label{fig:calibration-simplex}
\end{figure*}

To take into consideration  the effect of overconfident but wrong predictions, for a given dataset, we subtract the total counts of the cases when $\bm{y} \not= \bm{y}_{true}$ from the cases where $\bm{y} = \bm{y}_{true}$ for all $x_{test}$ in the dataset. Results are presented in the plots in Figure~\ref{fig:calibration-simplex} where we compare \adaicl against random and the best performing baseline on that dataset. For AGNews, TREC, and SST2, the label information of the retrieved examples is important and \adaicl leads to the best calibration in these cases. For RTE, MRPC, and MNLI, retrieving examples of the same label is not crucial for performance and thus, all methods behave similarly with respect to simplex calibration.

\subsection{Selection Time Complexity}

In Table~\ref{tab:time}, we compare competing approaches based on their computation time during their selection process (during downstream inference, their time complexity is the same). Random selection has zero cost. Vote$k$ and \adaicl ($T=1$) have the same cost, while the cost doubles for \adaicl ($T=2$). Nevertheless, hyper-parameter $T$ for \adaicl can be tuned depending on the desired runtime of the selection process.
\begin{table}[h]
\centering
\caption{Time complexity analysis with 5-shot ICL for different selection processes over 300 examples on a GeForce RTX 3090 (24GB GPU).}
\label{tab:time}%
\resizebox{0.8\linewidth}{!}{
\begin{threeparttable}
    \begin{tabular}{l|cc}
        \toprule
        & Embedding Computation (SBERT)  & Model Uncertainty Estimation (GPT-Neo)  \\
        \midrule
        \textbf{Amazon} \\
        Random & 0 secs & 0 secs \\ 
        Vote$k$ & $\approx$1 secs & 1 min \& 51 secs \\ 
        \adaicl ($T=1$) & $\approx$1 secs & 1 min \& 51 secs \\ 
        \adaicl ($T=2$) & $\approx$1 secs & $\approx$ 3 mins \& 42 secs \\ 
        \hline
        \textbf{AGNews} \\
        Random & 0 secs & 0 secs \\ 
        Vote$k$ & $\approx$0.5 secs & 3 mins \& 48 secs \\ 
        \adaicl ($T=1$) & $\approx$0.5 secs & 3 mins \& 48 secs \\ 
        \adaicl ($T=2$) & $\approx$0.5 secs & $\approx$ 7 mins \& 36 secs \\ 
        \bottomrule
    \end{tabular}%

    \end{threeparttable}
}

\end{table}%

\section{\adaicl Ablation Studies} \label{app:Abla-D}

\subsection{Graph Ablation Studies} \label{app:graph}
\begin{table}[h]
\centering
\caption{Graph ablation study using GPT-Neo based on hyper-parameters $m$, which controls the number of graph neighbors, and $l$, which controls whether we consider 1-hop or 2-hop sets. The value combinations of $m$ and $l$ are adjusted via~\Eqref{eq:heuristic}.}
\label{tab:graph-abla}%
\resizebox{0.6\linewidth}{!}{
\begin{threeparttable}
    \begin{tabular}{l|ccc}
        \toprule
        & AGNews & SST2 & Amazon \\
        \midrule
        Vote$k$ & $62.77_{\pm 4.82}$ & $73.69_{\pm 9.05}$ &  $75.13_{\pm 0.98}$ \\ 
        \hline
        \adaicl \\
        \; $m=15, l=1$ & $68.61_{\pm 1.02}$ & $79.42_{\pm 1.28}$ & $77.34_{\pm 2.73}$ \\
        \; $m=25, l=1$ & $68.74_{\pm 3.59}$ & $77.60_{\pm 4.20}$& $76.95_{\pm 3.86}$\\
        *$m=5, l=2$ & $70.95_{\pm 1.87}$ & $79.68_{\pm 1.77}$ & $77.73_{\pm 2.23}$\\
        \; $m=7, l=2$ & $69.13_{\pm 1.39}$ & $78.12_{\pm 2.61}$ & $74.99_{\pm 0.84}$ \\
        \hline
       \adaiclp ($T=2$) \\
       *$m=15, l=1$ & $69.39_{\pm 1.35}$ & $79.03_{\pm 2.47}$ & $77.08_{\pm 1.50}$ \\
       \; $m=25, l=1$ & $68.87_{\pm 3.26}$ & $77.73_{\pm 2.22}$& $73.17_{\pm 2.59}$\\
        \; $m=5, l=2$ & $70.43_{\pm 1.60}$ & $77.73_{\pm 1.15}$ & $76.43_{\pm 2.55}$\\
        \; $m=7, l=2$ & $70.60_{\pm 2.89}$ & $76.95_{\pm 4.69}$ & $77.21_{\pm 2.12}$ \\
       
        \bottomrule
    \end{tabular}%
    \begin{tablenotes}
        \item *Denotes the default value.
    \end{tablenotes}

    \end{threeparttable}
}

\end{table}%

Table~\ref{tab:graph-abla} shows an ablation study on our proposed heuristic rule of~\Eqref{eq:heuristic}. We select $m$ such that it lies near the boundaries of ~\Eqref{eq:heuristic}, depending whether we choose $l=1$ hop sets or $l=2$ hop sets. As Table~\ref{tab:graph-abla} shows, our heuristic rule is robust and achieves good results with four different combinations. In some cases, having smaller values of $m$ leads to slightly better results as it excludes neighbors with less semantic similarity. 

\begin{table}[h]
\centering
\caption{Graph ablation study for \adaicl using GPT-J with different graph construction approaches. }
\label{tab:graph-abla2}%
\resizebox{0.9\linewidth}{!}{
\begin{threeparttable}
    \begin{tabular}{l|ccccccc}
        \toprule
        & AGNews & TREC & SST2 & Amazon & RTE & MRPC & MNLI \\
        \midrule
        $m$-nn graph & $77.08_{\pm 1.11}$ & $53.38_{\pm  5.10}$ & $84.24 _{\pm 1.32}$ & $92.45_{\pm 1.50}$ & $56.90_{\pm  1.75}$ & $70.17_{\pm 1.72}$ & $40.36_{\pm 1.75}$ \\
        $\delta$-graph & $76.17_{\pm 3.45}$ & $50.51_{\pm 4.65}$ & $ 81.90_{\pm 2.48}$ & $88.80_{\pm 1.57}$ & $56.63_{\pm 3.23}$ & $ 68.75_{\pm 1.93}$ & $40.62_{\pm 2.08}$ \\
        
        \bottomrule
    \end{tabular}%

    \end{threeparttable}
}

\end{table}%

Furthermore, we also experiment with using a threshold-based graph ($\delta$-graph) instead of the $m$-nn graph. To determine threshold $\delta$, we compute the cosine similarity between all nodes and set $\delta$ such as each node has $m$ neighbors \emph{on average} (at the $m$-nn graph each nodes has exactly $m$ neighbors). As Table~\ref{tab:graph-abla2} shows, using the $\delta$-graph performs slightly worse than the $m$-nn graph. We hypothesize that using the $\delta$-graph gives more importance on the semantics of the train distribution (as $\delta$ value is computed based on the similarity scores between all train examples), which may not always generalize well to the test distribution.  

\subsection{Uncertainty Threshold} \label{app:threshold}

\begin{table}[h]
\centering
\caption{Ablation study using GPT-Neo based on hyper-parameter $\theta$, which controls the number of the examples that are considered as hard ones.}
\label{tab:thres-abla}%
\resizebox{0.6\linewidth}{!}{
\begin{threeparttable}
    \begin{tabular}{l|ccc}
        \toprule
        & TREC & SST2 & Amazon \\
        \midrule
        Vote$k$ & $53.12_{\pm 4.07}$ & $73.69_{\pm 9.05}$ &  $75.13_{\pm 0.98}$ \\ 
        \hline
        \adaiclp ($\theta=0.5$) & $59.89_{\pm 2.07}$ & $79.03_{\pm 2.47}$ & $77.08_{\pm 1.50}$ \\
        \adaiclp ($\theta=0.33$) & $60.28_{\pm 3.13}$ & $78.77_{\pm 2.59}$ & $	78.90_{\pm 1.14}$ \\
        
        \bottomrule
    \end{tabular}%

    \end{threeparttable}
}

\end{table}%

By default, we consider 50\% ($\theta=0.5$) of the examples with the lowest confidence as hard examples. Table~\ref{tab:thres-abla} shows results when we focus on harder examples by setting $\theta=0.33$ for \adaiclp. Interestingly, \adaiclp's performance can be further boosted with careful tuning of the uncertainty threshold. Thus, \emph{automatically} determining which examples are considered as hard examples for the models seems a promising research direction.

\begin{figure}[h]
  \centering
  \begin{subfigure}[b]{.8\linewidth}
  \includegraphics[width=1.0\linewidth]{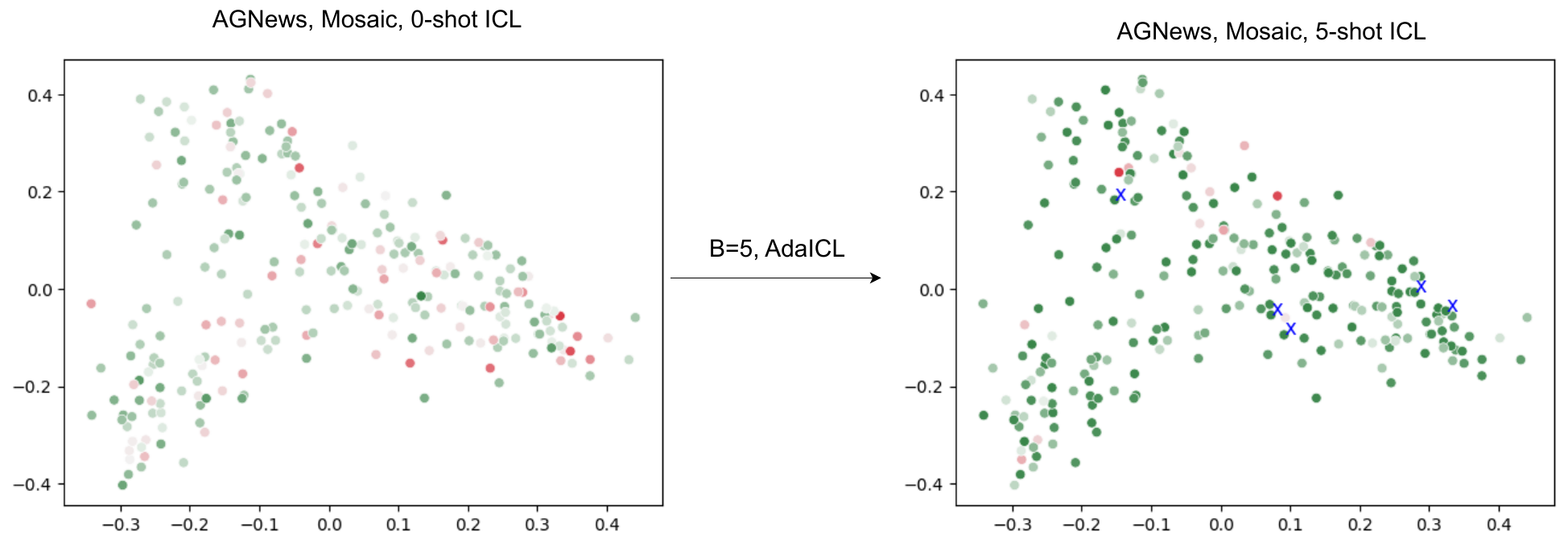}
    \caption{5-shot ICL by \adaicl selection for AGNews with Mosaic.}
    \vspace{0.1in}
  \end{subfigure}
  \begin{subfigure}[b]{.8\linewidth}
  \includegraphics[width=1.0\linewidth]{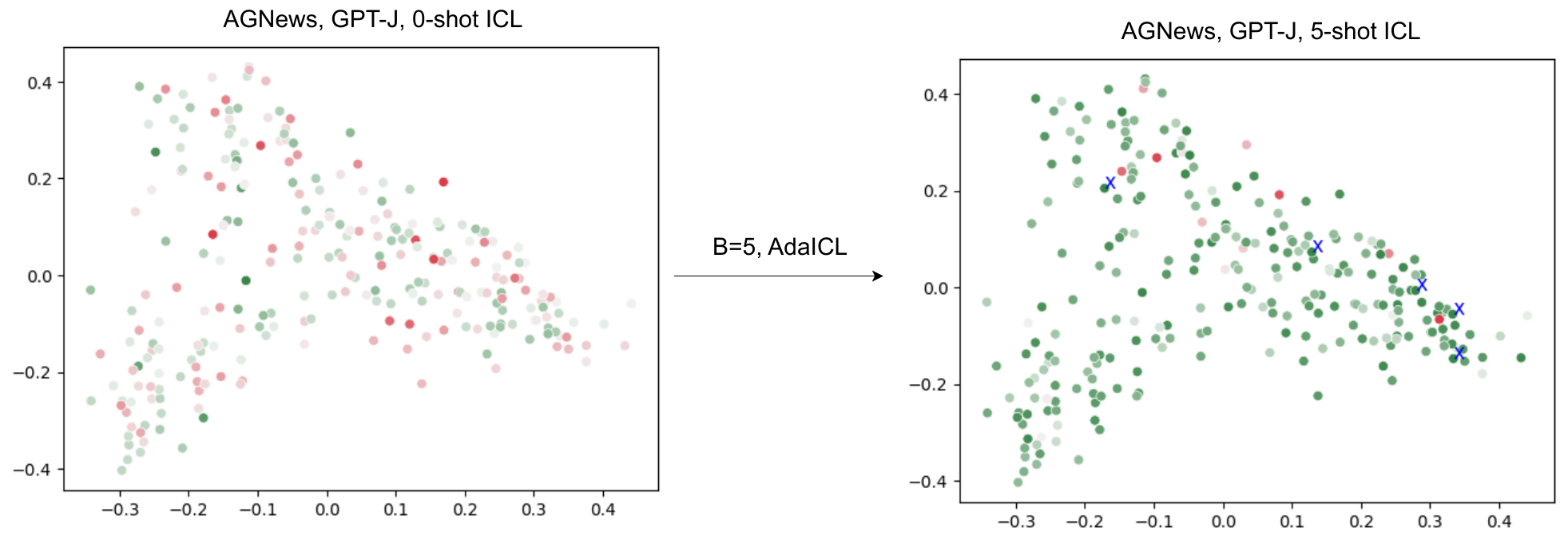}
    \caption{5-shot ICL by \adaicl selection for AGNews with GPT-J.}
  \end{subfigure}
  
  \begin{subfigure}[b]{.8\linewidth}
  \includegraphics[width=1.0\linewidth]{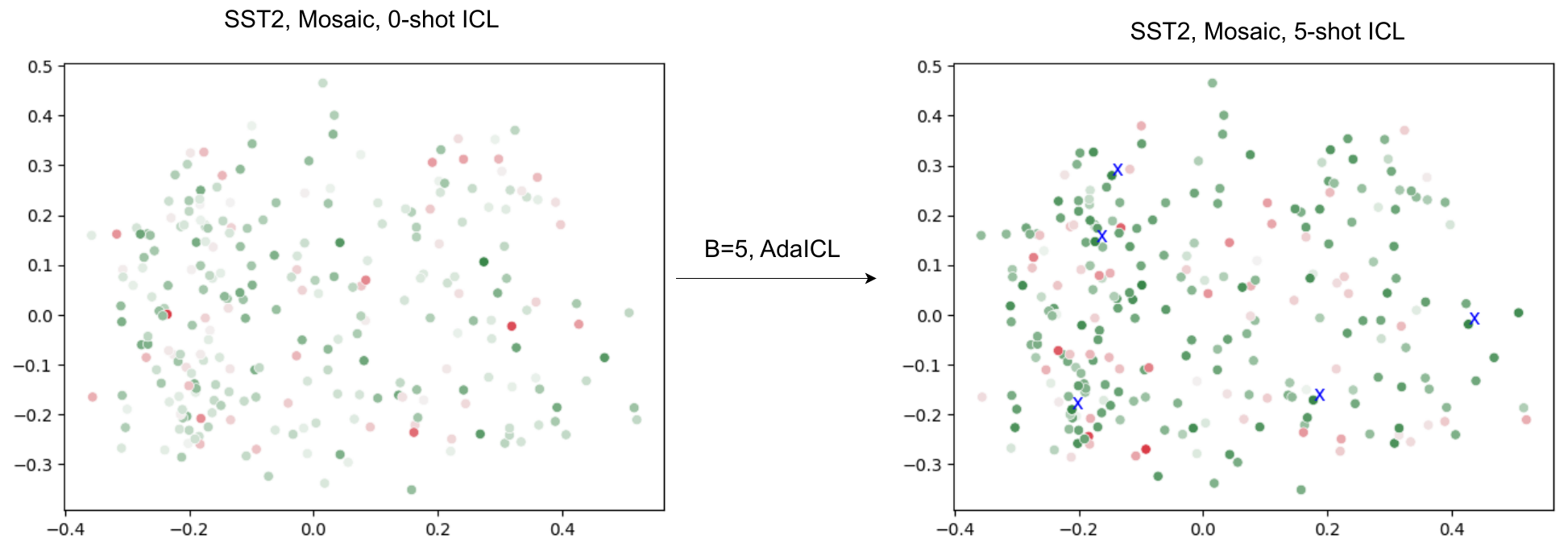}
    \caption{5-shot ICL by \adaicl selection for SST2 with Mosaic.}
  \end{subfigure}
  \begin{subfigure}[b]{.8\linewidth}
  \includegraphics[width=1.0\linewidth]{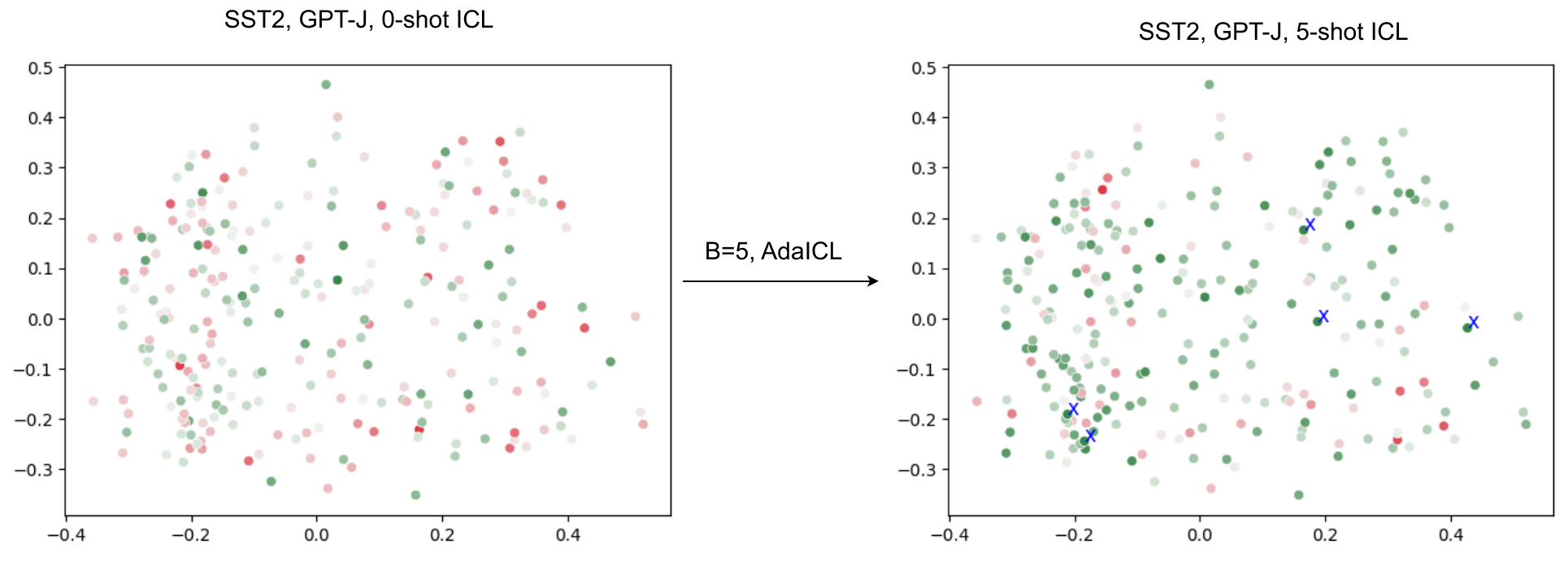}
    \caption{5-shot ICL by \adaicl selection for SST2 with GPT-J.}
  \end{subfigure}
  \caption{ Visualization of the SBERT space (after PCA) and of the examples selected for annotation. The two axes represent the two most important PCA components, which is performed over the SBERT embeddings of the examples. The hue color (green to red) represents the model's uncertainty (confident to uncertain) for each example. The selected examples for annotation are marked with the blue `X' symbol (better viewed with zooming). 
  }
    \label{fig:emb_plots}
\end{figure}

\section{Visualization} \label{app:visual}
We illustrate the selection process of \adaicl in~\Figref{fig:emb_plots}. Initially, the LLMs perform 0-shot ICL but do not make confident predictions (as the hue color,  that represents the model's uncertainty for each example, indicates). Note that different LLMs may consider different examples as hard or easy ones. Then, \adaicl selects 5 representative for 5-shot ICL, which improves the LLMs' understanding of the task and reduces its uncertainty (we observe fewer red nodes and more nodes with greener color).

\section{\adaicl Limitations}
We list some of our assumptions that may limit \adaicl if they are not satisfied. First, we assume that we can access the output logits/probabilities from the LLM in order to evaluate its uncertainty; this might not be always be feasible. Second, \adaicl relies on embedding methods to determine semantic diversity. While \adaicl is shown to be robust to different methods, it can still suffer if the semantic space of the test is wildly different from the annotation pool space. Finally, the graph/set construction is a heuristic approach and does not account for cases where adversarial examples are injected into the pool in order to degrade performance.

\end{document}